\documentclass{FrontiersinHarvard} % for articles in journals using the Harvard Referencing Style (Author-Date), for Frontiers Reference Styles by Journal: https://zendesk.frontiersin.org/hc/en-us/articles/360017860337-Frontiers-Reference-Styles-by-Journal
\usepackage{url,hyperref,lineno,microtype,subcaption}
\usepackage[onehalfspacing]{setspace}

\usepackage{times}
\usepackage{soul}
\usepackage{url}
\usepackage{hyperref}
\usepackage{inputenc}
\usepackage{caption}
\usepackage{graphicx}
\usepackage{amsmath}
\usepackage{amsthm}
\usepackage{booktabs}
\usepackage{algorithm}
\usepackage{algorithmic}
\usepackage{lineno}
\usepackage{nicefrac}
\usepackage{microtype}
\usepackage{multirow}
\usepackage{adjustbox}
\usepackage{array}
\usepackage{booktabs}
\usepackage{amsmath,amsfonts,amsthm,amssymb}
\usepackage{bm}
\usepackage{nicefrac}
\usepackage{microtype}
\usepackage{multirow}
\usepackage{graphicx}
\usepackage{enumerate}
\usepackage{titlesec}
\usepackage{listings}

\usepackage{titlesec}
\usepackage{listings}

\usepackage{makecell}
\usepackage{tabularx}
\usepackage{amsmath}
\usepackage{booktabs}

\usepackage{adjustbox}
\usepackage{array}
\usepackage{booktabs}
\usepackage{amsmath,amsfonts,amsthm,amssymb}
\usepackage{bm}
\usepackage{nicefrac}
\usepackage{microtype}
\usepackage{multirow}

\usepackage{color,xcolor}

\def\keyFont{\fontsize{8}{11}\helveticabold }
\def\firstAuthorLast{Guo {et~al.}} %use et al only if is more than 1 author
\def\Authors{Yufei Guo\,$^{1,2}$, Xuhui Huang\,$^{1,2}$, and Zhe Ma\,$^{1,2,*}$}
\def\Address{$^{1}$Intelligent Science \& Technology Academy of CASIC, Beijing, China \\
$^{2}$Scientific Research Laboratory of Aerospace Intelligent Systems and Technology, Beijing, China  }
\def\corrAuthor{Zhe Ma}

\def\corrEmail{mazhe\_thu@163.com}

\begin{document}
\onecolumn
%\firstpage{1}

\title {Direct Learning-Based Deep Spiking Neural Networks: A Review} 

\author[\firstAuthorLast ]{\Authors} %This field will be automatically populated
\address{} %This field will be automatically populated
\correspondance{} %This field will be automatically populated

\extraAuth{}% If there are more than 1 corresponding author, comment this line and uncomment the next one.
%\extraAuth{corresponding Author2 \\ Laboratory X2, Institute X2, Department X2, Organization X2, Street X2, City X2 , State XX2 (only USA, Canada and Australia), Zip Code2, X2 Country X2, email2@uni2.edu}

\maketitle

\begin{abstract}

%%% Leave the Abstract empty if your article does not require one, please see the Summary Table for full details.
\section{}
    The spiking neural network (SNN), as a promising brain-inspired computational model with binary spike information transmission mechanism, rich spatially-temporal dynamics, and event-driven characteristics, has received extensive attention. However, its intricately discontinuous spike mechanism brings difficulty to the optimization of the deep SNN. Since the surrogate gradient method can greatly mitigate the optimization difficulty and shows great potential in directly training deep SNNs, a variety of direct learning-based deep SNN works have been proposed and achieved satisfying progress in recent years. In this paper, we present a comprehensive survey of these direct learning-based deep SNN works, mainly categorized into accuracy improvement methods, efficiency improvement methods, and temporal dynamics utilization methods. In addition, we also divide these categorizations into finer granularities further to better organize and introduce them. Finally, the challenges and trends that may be faced in future research are prospected.

\tiny
 \keyFont{ \section{Keywords:} Spiking Neural Network, Brain-inspired Computation, Direct Learning, Deep Neural Network, Energy Efficiency, Spatial-temporal Processing} %All article types: you may provide up to 8 keywords; at least 5 are mandatory.
\end{abstract}

\section{Introduction}

The Spiking Neural Network (SNN) has been recognized as one of the brain-inspired neural networks due to its bio-mimicry of the brain neurons. It transmits information by firing binary spikes and can process the information in a spatial-temporal manner \citep{2020Incorporating,2018Direct,zhang2020efficient,wu2019deep,zhang2020supervised}. This event-driven and spatial-temporal manner makes the SNN very efficient and good at handling temporal signals, thus receiving a lot of research attention, especially recently.

Despite the energy efficiency and spatial-temporal processing advantages, it is a challenge to train deep SNNs due to the firing process of the SNN is undifferentiable, thus making it impossible to train SNNs via gradient-based optimization methods. At first, many works leverage the spike-timing-dependent plasticity (STDP) approach \citep{2020Spatial}, which is inspired by biology, to update the SNN weights. However, STDP cannot help train large-scale networks yet, thus limiting the practical applications of the SNN. There are two widely used effective pathways to obtain deep SNNs up to now. First, the ANN-SNN conversion approach \citep{2020Deep,li2021free,bu2022optimized,liu2022spikeconverter,li2022efficient,wang2022signed,bu2023optimal} converts a well-trained ANN to an SNN by replacing the activation function from ReLU with spiking activation. It provides a fast way to obtain an SNN. However, it is limited in the rate-coding scheme and ignores the rich temporal dynamic behaviors of SNNs. Second, the surrogate gradient (SG)-based direct learning approach \citep{2021Deep,2018Spatio,guo2022imloss,li2021differentiable} tries to find an alternative differentiable surrogate function to replace the undifferentiable firing activity when doing back-propagation of the spiking neurons. Since SG can handle temporal data and provide decent performance with few time-steps on the large-scale dataset, it has received more attention recently.

Considering the sufficient advantages and rapid development of the direct learning-based deep SNN, a comprehensive and systematic survey on this kind of work is essential. Previously related surveys \citep{wang2020supervised,zhang2022recent,roy2019towards,tavanaei2019deep,ponulak2011introduction,yamazaki2022spiking} have begun to classify existing works mainly based on the key components of SNNs: biological neurons, encoding methods, SNN structures, SNN learning mechanisms, software and hardware frameworks, datasets, and applications. Though such classification is intuitive to general readers, it is difficult for them to grasp the challenges and the landmark work involved. While in this survey, we provide a new perspective to summarize these related works, \textit{i.e.}, starting from analyzing the characteristics and difficulties of the SNN, and then classify them into i) accuracy improvement methods, ii) efficiency improvement methods, and iii) temporal dynamics utilization methods, based on the solutions for corresponding problems or the utilization of SNNs' advantages.

Further, these categories are divided into finer granularities: i) accuracy improvement methods are subdivided as improving representative capabilities and relieving training difficulties; ii) efficiency improvement methods are subdivided as network compression techniques and sparse SNNs; iii) temporal dynamics utilization methods are subdivided as sequential learning and cooperating with neuromorphic cameras. In addition to the classification by using strengths or overcoming weaknesses of SNNs, these recent methods can also be divided into the neuron level, network structure level, and training technique level, according to where these methods actually work.
%based on the modified or improved parts in SNNs. 
The classifications and main techniques of these methods are listed in \autoref{overview} and \autoref{overview2}. Finally, some promising future research directions are provided.

The organization of the remaining part is given as follows, \autoref{Preliminary} introduces the preliminary for spiking neural networks. The characteristics and difficulties of the SNN are also analyzed in \autoref{Preliminary}. \autoref{method} presents the recent advances falling into different categories. \autoref{Conclusions} points out future research trends and concludes the review.

\section{Preliminary}\label{Preliminary}
Since the neuron models are not the focus of the paper, here, we briefly introduce the commonly used discretized Leaky Integrate-and-Fire (LIF) spiking neurons to show the basic characteristic and difficulties in SNNs, which can be formulated by
\begin{equation}\label{eq:u}
	U^t_l= \tau U^{t-1}_l + \mathbf{W}_lO^t_{l-1},  \qquad  U^t_l < V_{\rm th},
\end{equation}
where $U^t_l$ is the membrane potential at $t$-th time-step for $l$-th layer, $O^t_{l-1}$ is the spike output from the previous layer, $\mathbf{W}_l$ is the weight matrix at $l$-th layer, $V_{th}$ is the firing threshold, and $\tau$ is a time leak constant for the membrane potential, which is in $\left( 0, 1 \right]$. When $\tau$ is $1$, the above equation will degenerate to the Integrate-and-Fire (IF) spiking neuron. 

\noindent\textbf{Characteristic 1.} \textit{Rich spatially-temporal dynamics. 
Seen from \autoref{eq:u}, different from ANNs, 
SNNs enjoy the unique spatial-temporal dynamic in the spiking neuron
model.}% ref?

Then, when the membrane potential exceeds the firing threshold, it will fire a spike and then fall to resting potential, given by
\begin{equation}\label{eq:ot}
	O^t_l=
	\left\{
	\begin{array}{lll}
		1, \quad {\rm{if}} \; U^t_l \ge V_{\rm th}\\
		0, \quad  \;  \rm{otherwise} \\
	\end{array}.
	\right.
\end{equation}
\noindent\textbf{Characteristic 2.} \textit{Efficiency. 
Since the output is a binary tensor, the multiplications of activations and weights can be replaced by additions, thus enjoying high energy efficiency. Furthermore, when there is no spike output generated, the neuron will keep silent. This event-driven mechanism can further save energy when implemented in neuromorphic hardware.}

\noindent\textbf{Characteristic 3.} \textit{Limited representative ability.
Obviously, transmitting information by quantizing the real-valued membrane potentials into binary output spikes will introduce the quantization error in SNNs, thus causing information loss~\citep{Guo2022eccv,wang2023mtsnn,guo2023rmploss,guo2022imloss}. Furthermore, the binary spike feature map from a timestep cannot carry enough information like the real-valued one in ANNs~\citep{guo2022real}}. These two problems limit the representative ability of SNN to some extent.

\noindent\textbf{Characteristic 4.} \textit{Non-differentiability. Another thorny problem in SNNs is the non-differentiability of the firing function.} 

To demonstrate this problem, we formulate the gradient at the layer $l$ by the chain rule, given by
\begin{equation}\label{eq:gradiet}
 \frac{\partial {L}}{\partial {\mathbf{W}_l}} = \sum_t (\frac{\partial {L}}{\partial {O^t_l}} \frac{\partial {O^t_l}}{\partial {U^t_l}} +  \frac{\partial {L}}{\partial {U^{t+1}_l}} \frac{\partial {U^{t+1}_l}}{\partial {U^{t}_l}} )\frac{\partial {U^{t}_l}}{\partial {\mathbf{W}_l}},
\end{equation}
where $\frac{\partial {O^t_l}}{\partial {U^t_l}}$ is the gradient of firing function at at $t$-th time-step for $l$-th layer and is $0$ almost everywhere, while infinity at $V_{\rm th}$. As a consequence, the gradient descent $(\mathbf{W}_l \leftarrow  \mathbf{W}_l - \eta \frac{\partial {L}}{\partial {\mathbf{W}_l}})$ either freezes or updates to infinity.

Most existing direct learning-based SNN works focus on solving difficulties or utilizing the advantages of SNNs. Boosting the representative ability and mitigating the non-differentiability can both improve SNN's accuracy. From this perspective, we organize the recent advances in the SNN field as accuracy improvement methods, efficiency improvement methods, and temporal dynamics utilization methods.

\section{Recent Advances}\label{method}

In recent years, a variety of direct learning-based deep spiking neural networks have been proposed. Most of these methods fall into solving or utilizing the intrinsic disadvantages or advantages of SNNs. Based on this, in the section, we classify these methods into accuracy improvement methods, efficiency improvement methods, and temporal dynamics utilization methods. In addition, these classifications are also organized in different aspects with a comprehensive analysis. \autoref{overview} and \autoref{overview2} summarizes the surveyed SNN methods in different categories.

Note that the direct learning methods can be divided into time-based methods and activation-based methods based on whether the gradient represents spike timing (time-based) or spike scale (activation-based)~\citep{zhu2022training}. In time-based methods, the gradients represent the direction where the timing of a spike should be moved, \textit{i.e.}, be moved leftward or rightward on the time axis. The SpikeProp~\citep{2002Error} and its variants~\citep{2005ABooij,2017Training,2013AXusupervised} all belong to this kind of method and they adopt the negative inverse of the time derivative of membrane potential function to approximate the derivative of spike timing to membrane potential. Since most of the time-based methods would restrict each neuron to fire at most once, in~\citep{zhou2021temporalcoded}, the spike time is directly taken  as the state of a neuron. Thus the relation of neurons can be modeled by the spike time and the SNN can be trained similarly to an ANN. Though the time-based methods enjoy less computation cost than the activation-based methods and many works~\citep{zhu2022training,zhang2021temporal} have greatly improved the accuracy of the field, it is still difficult to train deep time-based SNN models and apply them to large-scale datasets, \textit{e.g.}, ImageNet. Considering the limits of the time-based methods and the topic of summarizing the recent deep SNNs here, we mainly focus on activation-based methods in the paper.

\begin{table*}[tp]
\caption{Overview of Direct Learning-Based Deep Spiking Neural Networks: Part I. }
\label{overview}
\begin{center}
 \resizebox{\textwidth}{!}{
\begin{tabular}{ccllccc}
\toprule
\multicolumn{2}{c}{\multirow{2}{*}{\bf Type}}   & \multirow{2}{*}{\bf Method} & \multirow{2}{*}{\bf Key Technology} & \multicolumn{3}{c}{\bf On the Level$^\star$}\\
\cmidrule{5-7}
        &         &          &           &    NL  &   NSL  &  TTL \\
\midrule     

\multirow{54}{*}{\makecell[c]{Accuracy \\Improvement}} & \multirow{32}{*}{\makecell[c]{Improving \\ representative \\capabilities}} & LSNN~\citep{2018Long} & Adaptive threshold & \checkmark & &\\
    &   &  LTMD\citep{wang2022ltmd} & Adaptive threshold & \checkmark & & \\
    &   &  BDETT~\citep{ding2022biologically} &  Dynamic threshold & \checkmark & & \\ 
    &   &  PLIF~\citep{2020Incorporating} &  Learnable leak constant & \checkmark & & \\ 
    &   &  Plastic Synaptic Delays~\citep{ImprovingYu2022} &  Learnable leak constant & \checkmark & & \\ 
    &   &   Diet-SNN~\citep{2020DIET} &  Learnable leak constant\& threshold & \checkmark & & \\ 
    &   &   DS-ResNet~\citep{FengLTM022} &  Multi-firing \& Act before Add-ResNet & \checkmark & \checkmark & \\ 
    &   &   SNN-MLP~\citep{2022Brain} &   Group LIF & \checkmark & & \\ 
    &   &   GLIF~\citep{GLIF} &   Unified gated LIF & \checkmark & & \\ 
    &   &   Augmented Spikes~\citep{SynapticYu2022} &   Augmented spikes & \checkmark & & \\ 
    &   &   InfLoR-SNN~\citep{shen2023exploiting} &   Leaky Integrate and Fire or Burst  & \checkmark & & \\ 
    &   &   MT-SNN ~\citep{wang2023mtsnn} &   Multiple threshold approach & \checkmark & & \\ 

    &   &   SEW-ResNet~\citep{2021Deep} &   Act before ADD form-based ResNet &  & \checkmark & \\ 
    &   &   MS-ResNet~\citep{2021Advancing} &   Pre-activation form-based ResNet &  & \checkmark & \\ 
    &   &   AutoSNN~\citep{na2022autosnn} &    Neural architecture search &  & \checkmark & \\ 
    &   &   SNASNet~\citep{Kim2022Neural} &    Neural architecture search &  & \checkmark & \\ 
    &   &   TA-SNN~\citep{yao2021temporal} &   Attention mechanism &  & \checkmark & \\
    &   &   STSC-SNN~\citep{yu2022stsc} &   Attention mechanism &  & \checkmark & \\ 
    &   &   TCJA-SNN~\citep{zhu2022tcjasnn} &   Attention mechanism &  & \checkmark & \\
    &   &   Spikformer~\citep{zhou2022spikformer} &   Transformer-based SNN &  & \checkmark & \\
    &   &   Spikingformer~\citep{zhou2023spikingformer} &   Transformer-based SNN &  & \checkmark & \\
    &   &   Auto-Spikformer~\citep{che2023autospikformer} &   Transformer-based SNN &  & \checkmark & \\
    &   &    Real Spike~\citep{guo2022real} &   Training-inference decoupled structure &  & \checkmark & \\
    &   &    IM-Loss~\citep{guo2022imloss} &   Information maximization loss &  &  & \checkmark\\
    &   &    RecDis-SNN~\citep{Guo_2022_CVPR} & Membrane potential distribution loss &  &  & \checkmark\\
     &   &   RMP-Loss~\citep{guo2023rmploss} & Membrane potential distribution loss&  &  & \checkmark\\    
    &   &    Distilling spikes~\citep{kushawaha2021distilling} & Knowledge distillation &  &\checkmark  & \checkmark\\
    &   &    Local Tandem Learning ~\citep{yang2022training} & Tandem Learning &  &  & \checkmark\\ 
    &   &    sparse-KD~\citep{xu2023biologically} & Knowledge distillation &  &  & \checkmark\\
    &   &    KDSNN~\citep{xu2023constructing} & Knowledge distillation &  &  & \checkmark\\    
    &   &    SNN distillation~\citep{takuya2021training} & Knowledge distillation &  &  & \checkmark\\

\cmidrule{2-7}
    & \multirow{22}{*}{\makecell[c]{Relieving \\ training \\difficulties}}  &    SuperSpike~\citep{2017SuperSpike} & Fixed surrogate gradient &  &  & \checkmark\\  
    &   &    LISNN~\citep{2020LISNN} & Fixed surrogate gradient &  &  & \checkmark\\
    &   &    IM-Loss~\citep{guo2022imloss} & Dynamic surrogate gradient &  &  & \checkmark\\    
    &   &    Gradual surrogate gradient~\citep{guo2022imloss} & Dynamic surrogate gradient &  &  & \checkmark\\ 
    &   &    Differentiable Spike~\citep{li2021differentiable} & Learnable surrogate gradient &  &  & \checkmark\\  
    &   &    SpikeDHS~\citep{leng2022differentiable} & Differentiable surrogate gradient search &  &  & \checkmark\\  
   &   &   DSR~\citep{meng2022training} & Differentiation on Spike Representation &  &  & \checkmark\\ 
      &   &   NSNN~\citep{ma2023exploiting} & Noise-driven learning rule &  &  & \checkmark\\ 
    &   &    STDBP~\citep{ZhangRectified2022} & Rectified postsynaptic potential function &  & \checkmark & \\ 
    &   &    SEW-ResNet~\citep{2021Deep} & Act before ADD form-based ResNet &  & \checkmark & \\ 
    &   &    MS-ResNet~\citep{2021Advancing} & Pre-activation form-based ResNet &  & \checkmark & \\   
     &   &   NeuNorm~\citep{2018Direct} & Constructing auxiliary feature maps&  &  & \checkmark\\    
     &   &   tdBN~\citep{2020Going} & Threshold-dependent batch normalization&  &  & \checkmark\\  
     &   &   BNTT~\citep{2020Revisiting} & Temporal batch normalization through time&  &  & \checkmark\\ 
     &   &   PSP-BN~\citep{Rethinking2022} & Postsynaptic potential normalization&  & \checkmark & \\ 
     &   &   TEBN~\citep{2020Revisiting} & Temporal effective batch normalization&  &  & \checkmark\\ 
     &   &   MPBN~\citep{guo2023membrane} & Membrane potential batch normalization&  &  & \checkmark\\ 
     &   &   RecDis-SNN~\citep{Guo_2022_CVPR} & Membrane potential distribution loss&  &  & \checkmark\\
     &   &   TET~\citep{deng2022temporal} & Temporal regularization loss&  &  & \checkmark\\ 
     &   &   Tandem learning~\citep{2021ATandem} & Tandem learning&  &  & \checkmark\\      
     &   &  Progressive tandem learning~\citep{2020Progressive} & Progressive tandem learning&  &  & \checkmark\\    
     &   &  Joint A-SNN~\citep{guo2023joint} & Joint training of ANN and SNN&  &  & \checkmark\\      
\bottomrule
\end{tabular}}
%\end{tabular}
\end{center}
{$^\star$ NL denotes Neuron Level, NSL denotes Network Structure Level, TTL denotes Training Technique Level.}
\end{table*}

\subsection{Accuracy Improvement Methods}

As aforementioned, the limited information capacity and the non-differentiability of firing activity of the SNN cause its accuracy loss for wide tasks. Therefore, to mitigate the accuracy loss in the SNN, a great number of methods devoted to improving the representative capabilities and relief training difficulties of SNNs have been proposed and achieved successful improvements in the past few years.

\subsubsection{Improving representative capabilities}

Two problems result in the representative ability decreasing of the SNN, the process of firing activity will induce information loss, which has been proved in~\citep{Guo2022eccv} and binary spike maps suffer the limited information capacity, which has been proved in~\citep{guo2022real}. These problems can be mitigated on the neuron level, network structure level, and training technique level.

\textbf{On the neuron level.} A common way to boost the representative capability of the SNN is to make some hyper-parameters in the spiking neuron learnable.
In LSNN~\citep{2018Long} and LTMD~\citep{wang2022ltmd}, the adaptive threshold spike neuron was proposed to enhance the computing and learning capabilities of SNNs.
Further, a novel bio-inspired dynamic energy-temporal threshold, which can be adjusted dynamically according to input data for SNNs was introduced in the BDETT~\citep{ding2022biologically}.
%Following the same idea to set some hyper-parameters in the spike neuron be learnable, 
Some works adopted the learnable membrane time constant in spiking neurons~\citep{2020Effective,2019Technical,2020Incorporating,SupervisedLuo2022,ImprovingYu2022}.
Combining these two manners, Diet-SNN~\citep{2020DIET} simultaneously adopted the learnable membrane leak and firing threshold.

There are also some works focusing on embedding more factors in the spiking neuron to improve its diversity. A multi-level firing (MLF) unit, which contains multiple
LIF neurons with different level thresholds thus could generate more quantization spikes with different thresholds was proposed in DS-ResNet~\citep{FengLTM022}. A full-precision LIF to communicate between patches in Multi-Layer Perceptron (MLP), including horizontal LIF and vertical LIF in different directions was proposed in SNN-MLP~\citep{2022Brain}. SNN-MLP used group LIF to extract better local features.
In GLIF~\citep{GLIF}, to enlarge the representation space of spiking neurons, a unified gated leaky integrate-and-fire Neuron was proposed to fuse different bio-features in different neuronal behaviors via embedding gating factors. In augmented spikes~\citep{SynapticYu2022}, a special spiking neuron model was proposed to process augmented spikes, where additional information can be carried from spike strength and latency. This neuron model extends the computation with an additional dimension and thus could be of great significance for the representative ability of the SNN. In LIFB~\citep{shen2023exploiting}, a new spiking neuron model called the Leaky Integrate and Fire or Burst was proposed. The neuron model exhibits three modes including resting, regular spike, and burst spike, which significantly enriches the representative capability. Similar to LIFB, MT-SNN~\citep{wang2023mtsnn}  proposed a multiple threshold approach to firing different spike modes to alleviate the quantization error, such that it could reach a high accuracy at fewer steps.

Different from these works, InfLoR-SNN~\citep{Guo2022eccv} proposed a membrane potential rectifier (MPR), which can adjust the membrane potential to a new value closer to quantization spikes than itself before firing activity. MPR directly handles the quantization error problem in SNNs, thus improving the representative ability.

\begin{figure}
    \centering
    \includegraphics[width=0.6\linewidth]{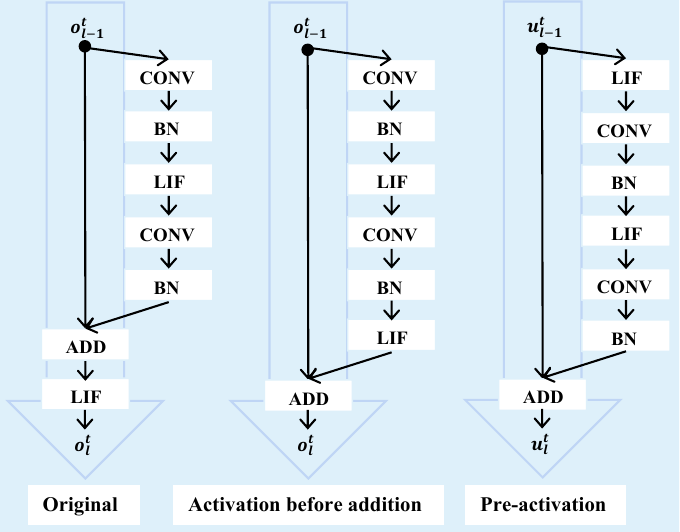}
    \caption{Different SNN ResNet architectures. }
    \label{struct}
\end{figure}

\textbf{On the network structure level.} To increase the SNN diversity, some works advocate for improving the SNN architecture. In SEW-ResNet~\citep{2021Deep} and DS-ResNet~\citep{FengLTM022}, the widely used standard ResNet backbone is replaced by activation before addition form-based ResNet. In this way, the blocks in the network will fire positive integer spikes. Its representation capability will no doubt be increased, however, the advantages of event-driven and multiplication-addition transform in SNNs will be lost in the meantime. To solve the aforementioned problem, MS-ResNet~\citep{2021Advancing} adopted the pre-activation form-based ResNet. In this way, the spike-based convolution can be retained. The difference between these methods is shown in~\autoref{struct}. However, these SNN architectures are all manually designed. For designing well-performed SNN models automatically, AutoSNN~\citep{na2022autosnn} and SNASNet~\citep{Kim2022Neural} combined the Neural Architecture Search (NAS) approach to find better SNN architectures. And TA-SNN~\citep{yao2021temporal}, STSC-SNN~\citep{yu2022stsc}, and TCJA-SNN~\citep{zhu2022tcjasnn} leveraged the learnable attention mechanism to improve the SNN performance. Spikformer~\citep{zhou2022spikformer}, Spikingformer~\citep{zhou2023spikingformer}, and Auto-Spikformer~\citep{che2023autospikformer} proposed the Transformer-based SNN.

Different from changing the network topology, Real Spike~\citep{guo2022real} provides a training-inference decoupled structure. This method enhances the representation capacity of the SNN by learning real-valued spikes during training. While in the inference phase, the rich representation capacity will be transferred from spike neurons to the convolutions by a re-parameterization technique, and meanwhile, the real-valued spikes will be transformed into binary spikes, thus maintaining the event-driven and multiplication-addition transform advantages of SNNs.

Besides, increasing the timestep of SNN will undoubtedly improve the SNN accuracy too, which has been proved in many works~\citep{2021Deep,2018Spatio,2018Direct}. To some extent, increasing the timestep is equivalent to increasing neuron output bits through the temporal dimension, which will increase the representation capability of feature map~\citep{FengLTM022}. However, using more timesteps achieves better performance at the cost of increasing inference time.
%However, we don't think increasing the time-step is a good strategy, due to that this will increase energy consumption of the SNN, which is contrary to the original intention of SNN to reduce energy cost.

\textbf{On the training technique level.} Some works attempted to improve the representative capability of the SNN on the training technique level, which can be categorized as regularization and distillation. Regularization is a technique that introduces another loss term to explicitly regularize the membrane potential or spike distribution to retain more useful information in the network that could indirectly help train the network as follows,
\begin{equation}\label{loss}
	\mathcal{L}_{Total} = \mathcal{L}_{CE}+ \lambda \mathcal{L}_{DL}
\end{equation}
where $\mathcal{L}_{CE}$ is the common cross-entropy loss, $\mathcal{L}_{DL}$ is the distribution loss for learning the proper membrane potential or spike, and $\lambda$ is a coefficient to balance the effect of the two types of losses. IM-Loss~\citep{guo2022imloss} argues that improving the activation information entropy can reduce the quantization error, and proposed an information maximization loss function that can maximize the activation information entropy. In RecDis-SNN~\citep{Guo_2022_CVPR}, a loss for membrane potential distribution to explicitly penalize three undesired shifts was proposed. Though the work is not designed for reducing quantization error specifically, it still results in a bimodal membrane potential distribution, which has been proven can mitigate the quantization error problem. RMP-Loss~\citep{guo2023rmploss} proposes a regularizing membrane potential loss (RMP-Loss) to force the membrane potential close to the spikes, which is directly related to quantization error. 

The distillation methodology aims to help train a small student model by transferring knowledge of a rather large trained teacher model based on the consensus that the representative ability of a teacher model is better than that of the student model. Recently, some interesting works that introduce the distillation method in the SNN domain were proposed. In~\citep{kushawaha2021distilling}, a big teacher SNN model is used to guide the small SNN counterpart learning. While in~\citep{yang2022training,takuya2021training,xu2023constructing,xu2023biologically}, an ANN-teacher is used to guide SNN-student learning. In specific, Local Tandem Learning ~\citep{yang2022training} uses the intermediate feature representations of the ANN to supervise the learning of SNN. While in sparse-KD~\citep{xu2023biologically}, the logit output of the ANN was adopted to guide the learning of the SNN. Furthermore, KDSNN~\citep{xu2023constructing} and SNN distillation~\citep{takuya2021training} used both feature-based and logit-based information to distill the SNN.
%How to use an ANN model to guide the training of an SNN model is remained study.

\subsubsection{Relieving training difficulties}

The non-differentiability of the firing function impedes the deep SNN direct training. To handle this problem, recently, using the surrogate gradient (SG) function for spiking neurons has received much attention. SG method utilizes a differentiable surrogate function to replace the non-differentiable firing activity to calculate the gradient in the back-propagation~\citep{2021Deep,2020DIET,2018Direct,2019Surrogate}. Though the SG method can alleviate the non-differentiability problem, there exists an obvious gradient mismatch between the gradient of the firing function and the surrogate gradient. And the problem easily leads to under-optimized SNNs with severe performance degradation. Intuitively, an elaborately designed surrogate gradient can help to relieve the gradient mismatch in the backward propagation. As a consequence, some works are focusing on designing better surrogate gradients. In addition, the gradient explosion/vanishing problem in SNNs is severer over ANNs, due to the adoption of \texttt{tanh}-like function for most SG methods. There are also some works focusing on handling the gradient explosion/vanishing problem. Note that, these methods in this section can also be classified as the improvement on the neuron level, network structure level, and training technique level, which can be seen in the \autoref{overview}. Nevertheless, to better introduce these works, we still organize them as designing the better surrogate gradient and relieving the gradient explosion/vanishing problem.
% However, here we organize them as designing the better surrogate gradient and relieving the gradient explosion/vanishing problem to better organize and introduce recent works.

\begin{figure}[t]
	\centering
	\includegraphics[width=0.4\textwidth]{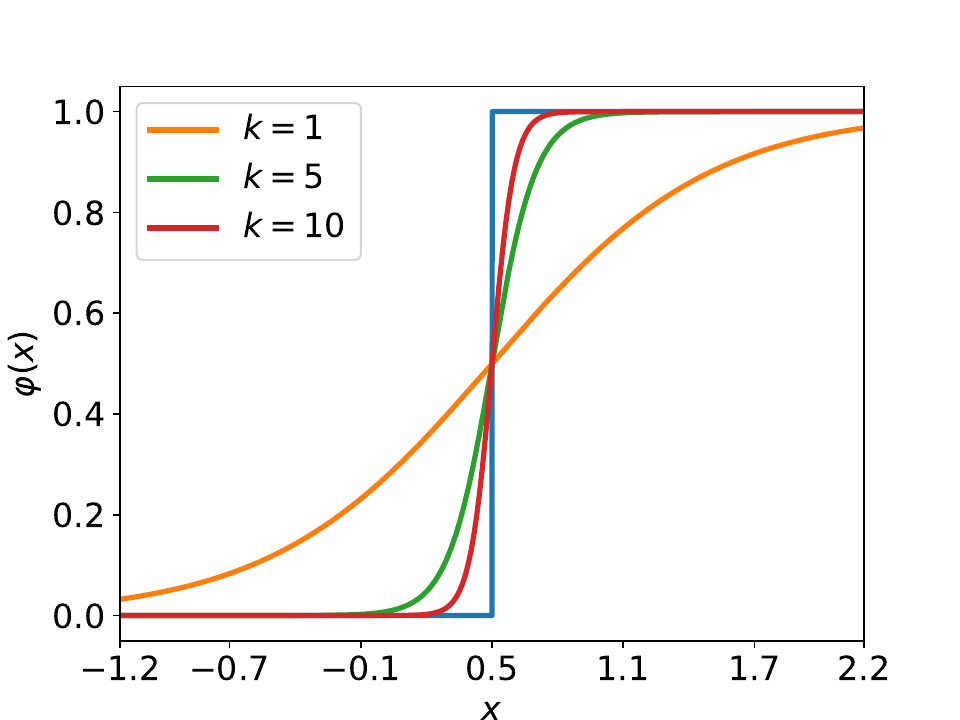}
	%	\hspace{0.1in}
	\includegraphics[width=0.4\textwidth]{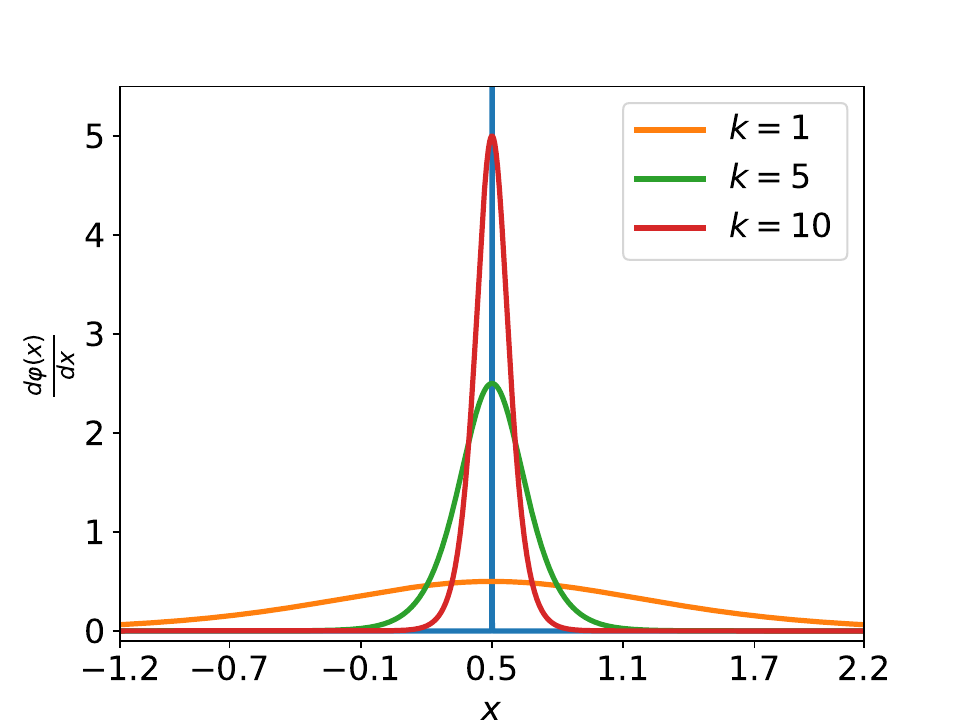}
	\caption{ The approximation function (left) under different values of the coefficient, $k$ and its corresponding gradient (right). The blue curves represent the firing function (left) and its true gradient (right).}
	\label{esg}
\end{figure}

\textbf{Designing the better surrogate gradient (SG)}. Most earlier works adopt fixed SG-based methods to handle the non-differentiability problem. For example, the derivative of a truncated quadratic function, the derivatives of a sigmoid, and a rectangular function were respectively adopted in ~\citep{2011Error}, \citep{2017SuperSpike}, and \citep{2020LISNN}. However, such a strategy would limit the learning capacity of the network. To this end, a dynamic SG method was proposed in \citep{guo2022imloss,ChenGradual2022}, where the SG could change along with epochs as follows,
\begin{equation}\label{evaf}
	\varphi (x) = \frac{1}{2} {\rm tanh}(K(i)(x-V_{\rm th}))+\frac{1}{2}
\end{equation}
where $\varphi (x)$ is the backward approximation function for the firing activity and $K(i)$ is a dynamic coefficient that changes along with the training epoch as follows,
\begin{equation}\label{evok}
	K(i) = \frac{(10^{\frac{i}{N}}-10^0)K_{\rm max}+ (10^1-10^{\frac{i}{N}})K_{\rm min} }{9} 
\end{equation}
where $K_{\rm min}$ and $K_{\rm max}$ are the lower bound and the upper bound of $K$, and $i$ is the index of epoch starting from $0$ to $N-1$. The $\varphi (x)$ and its gradient can be seen in~\autoref{esg}. Driven by $K(i)$, it will gradually evolve to the firing function, thus ensuring sufficient weight updates at the beginning and accurate gradients at the end of the training. Nevertheless, the above SG methods are still designed manually. To find the optimal solution, in \citep{li2021differentiable}, the Differentiable Spike method that can adaptively evolve during training to find the optimal shape and smoothness for gradient estimation based on the finite difference technique was proposed. Then, in~\citep{leng2022differentiable}, combined with the NAS technique, a differentiable SG search (DGS) method to find the optimized SGs for SNN was proposed. Different from designing a better SG for firing function, DSR~\citep{meng2022training} derived that the spiking dynamics with spiking neural models can be represented as some sub-differentiable mapping and trained the SNNs by the gradients of the mapping, thus avoiding the non-differentiability problem in SNN training. And NSNN~\citep{ma2023exploiting} presented the noisy spiking neural network and the noise-driven learning rule (NDL) for the surrogate gradient.
%The spatio-temporal back-propagation (STBP)~\citep{2018Spatio} method enables SNNs to be trained on the ANN programming platform, which also significantly promotes the SG methods.

\textbf{Relieving the gradient explosion/vanishing problem}.
The gradient explosion or vanishing problem is still severe in SG-only methods. There are three kinds of methods to solve this problem: using improved neurons or architectures, improved batch normalizations, and regularization. In~\citep{ZhangRectified2022}, a simple yet efficient rectified linear postsynaptic potential function (ReL-PSP) for spiking neurons, which benefits for handling the  gradient explosion problem, was proposed. On the network architecture level,
SEW-ResNet~\citep{2021Deep} showed that standard spiking ResNet is inapplicable to overcome identity mapping and vanishing/explosion gradient problems and advised using ResNet with activation before addition form. 
%As aforementioned, this modification will lose the multiplication-addition transform and event-driven advantages of SNNs. 
Recently, the pre-activation form-based ResNet was explored in MS-ResNet~\citep{2021Advancing}. This network topology can simultaneously handle the gradient explosion/vanishing problem and retain the advantages of the SNN.

The normalization approaches are widely used in ANNs to train well-performed models, and these approaches are also introduced in the SNN field to handle the vanishing/explosion gradient problems. For example, NeuNorm~\citep{2018Direct} normalized the data along the channel dimension like BN in ANNs through constructing auxiliary feature maps. Threshold-dependent batch normalization (tdBN)~\citep{2020Going} considers the SNN normalization from a temporal perspective and extends the scope of BN to the additional temporal dimension. Furthermore, some works~\citep{2020Revisiting,Rethinking2022,duan2022temporal} argued that the distributions of different timesteps vary wildly, thus bringing a negative impact when using shared parameters. Subsequently, the temporal Batch Normalization Through Time (BNTT), postsynaptic potential normalization (PSP-BN), and temporal effective batch normalization (TEBN) that can regulate the spike flows by utilizing separate sets of BN parameters on different timesteps were proposed.
Though adopting temporal BN parameters on different timesteps can obtain more well-performed SNN models, this kind of BN technique can not fold the BN parameters into the weights and will increase the computations and running time in the inference stage. Nevertheless, all these BNs above are advised to be used after convolution layers. However, the data flow
after the convolution layer will not be presented to the firing function directly but to the membrane potential updating function first. Hence, the data flow will be disturbed again before reaching the firing function. To handle the problem, MPBN~\citep{guo2023membrane} proposed to add another BN after the membrane potential updating function to retain normalized data flow before the firing function. It also provided a training-inference-decoupled re-parameterization technique to fold the trained MPBN into the firing threshold to eliminate the extra time burden in- duced by MPBN in the inference time.

Using the regularization loss can also mitigate the gradient explosion/vanishing problem. In RecDis-SNN~\citep{Guo_2022_CVPR}, a new perspective to further classify the gradient explosion/vanishing difficulty of SNNs into three undesired shifts of the membrane potential distribution was presented. To avoid these undesired shifts, a membrane potential regularization loss was proposed in RecDis-SNN, this loss introduces no additional operations in the SNN inference phase. In TET~\citep{deng2022temporal}, an extra temporal regularization loss
to compensate for the loss of momentum in the gradient descent with SG methods was proposed. With this loss, TET can converge into flatter minima with better generalizability.

Since ANNs are fully differentiable to be trained with gradient descent, there is also some work utilizing ANN to guide the SNN’s optimization~\citep{2020Progressive,2021ATandem,guo2023joint}. In~\citep{2021ATandem} a tandem learning framework was proposed,
that consists of an SNN and an ANN that share the same weight. In this framework, the spike count as the discrete neural representation in
the SNN would be presented to the coupled ANN  activation function in the forward phase. And in the backward phase,  the error back-propagation
is performed on the ANN to update the shared weight for both the SNN and the ANN. Furthermore, in ~\citep{2020Progressive}, a progressive tandem learning framework was proposed, that introduces a layer-wise learning method to fine-tune the the shared network weights. Considering the difference between the ANN and SNN, Joint A-SNN~\citep{guo2023joint}  developed a partial weight-sharing regime for the joint training of weight-shared ANN and SNN, that applies the Singular Value
Decomposition (SVD) to the weights parameters and keep the same eigenvectors while the separated eigenvalues for the ANN and SNN.

\begin{table*}[tp]
\caption{Overview of Direct Learning-Based Deep Spiking Neural Networks: Part II }
\label{overview2}
\begin{center}
 \resizebox{\textwidth}{!}{
\begin{tabular}{ccllccc}
\toprule
\multicolumn{2}{c}{\multirow{2}{*}{\bf Type}}   & \multirow{2}{*}{\bf Method} & \multirow{2}{*}{\bf Key Technology} & \multicolumn{3}{c}{\bf On the Level$^\star$}\\
\cmidrule{5-7}
        &         &          &           &    NL  &   NSL  &  TTL \\
\midrule     

\multirow{17}{*}{\makecell[c]{Efficiency \\Improvement}} & \multirow{12}{*}{\makecell[c]{Network \\ compression \\techniques}} & Spatio-Temporal Pruning~\citep{2021Spatio} & Spatio-temporal pruning &  & &\checkmark\\
    &   &  SD-SNN~\citep{han2022adaptive} & Pruning-regeneration method &  & & \checkmark\\
    &   &  Grad R ~\citep{chen2021pruning} & Pruning-regeneration method &  & & \checkmark\\
    &   &  Temporal pruning ~\citep{Chowdhury2022} & Temporal pruning &  & & \checkmark\\
    &   &  Autosnn ~\citep{na2022autosnn} & Neural architecture searching &  &\checkmark & \\
    &   &  SNASNet ~\citep{Kim2022Neural} & Neural architecture searching &  &\checkmark & \\
    &   &  Lottery Ticket Hypothesis ~\citep{kim2022exploring} & Lottery ticket hypothesis &  & \checkmark& \\
    &   &    Distilling spikes~\citep{kushawaha2021distilling} & Knowledge distillation &  &\checkmark  & \checkmark\\
    &   &    Local Tandem Learning ~\citep{yang2022training} & Tandem Learning &  &  & \checkmark\\ 
    &   &    sparse-KD~\citep{xu2023biologically} & Knowledge distillation &  &  & \checkmark\\
    &   &    KDSNN~\citep{xu2023constructing} & Knowledge distillation &  &  & \checkmark\\    
    &   &    SNN distillation~\citep{takuya2021training} & Knowledge distillation &  &  & \checkmark\\   
\cmidrule{2-7}
    & \multirow{5}{*}{\makecell[c]{Sparse \\ SNNs}}  &    ASNN~\citep{zambrano2016fast} & A lot of adaptive spiking neurons & \checkmark &  & \\  
    &   &    Correlation-based regularization~\citep{HanCorrelation2022} & Correlation-based regularizer &  &  & \checkmark\\
    &   &   Superspike~\citep{2017SuperSpike} & Heterosynaptic regularization term &  &  & \checkmark\\
    &   &   RecDis-SNN~\citep{Guo_2022_CVPR} &  Membrane potential distribution &  &  & \checkmark\\
    &   &   Low-activity SNN~\citep{Pellegrini2021Low} &   Regularization
term &  & \checkmark & \checkmark\\
\midrule 
\multirow{22}{*}{\makecell[c]{Temporal \\Dynamics \\ Utilization}} & \multirow{10}{*}{\makecell[c]{Sequential \\ learning }} & Sequence approximation~\citep{she2021sequence} &  Dual-search-space optimization &  & &\checkmark\\
    &   &  Sequential learning~\citep{ponghiran2022spiking} & Improved recurrence dynamics & \checkmark & & \\
    &   &  SNN\_HAR~\citep{li2022wearable} & Spatio-temporal extraction &  &\checkmark & \\
    &   &  Robust SNN~\citep{NomuraRobustness2022} & Temporal penalty settings &  &\checkmark & \\
    &   &  Tandem learning-based SNN model~\citep{articleWu2020} & Tandem learning &  & & \checkmark\\
    &   &  SG-based SNN model~\citep{bittar2023surrogate} & Surrogate gradient method &  & & \checkmark\\ 
    &   &  Combination-based SNN~\citep{articleBittar2022} & Combination of many techniques& \checkmark & \checkmark & \\   
    &   &   Low-activity SNN~\citep{Pellegrini2021Low} &   Regularization
term &  &  & \checkmark\\
    &   &   SNNCNN~\citep{Sadovsky2023Speech} &   Combination of CNNs and SNNs &  & \checkmark &\checkmark \\ 
    &   &   RSNNs~\citep{articleYin2021} &   activity-regularizing SG &  & \checkmark &\checkmark \\     
\cmidrule{2-7}
    & \multirow{12}{*}{\makecell[c]{Cooperating \\ with\\neuromorphic\\cameras}}  &   daptive-spikenet~\citep{kosta2022adaptive} & Learnable neuronal dynamics& \checkmark &  & \\  
    &   &   StereoSpike~\citep{ranccon2021stereospike} & Modified U-Net-like  architecture &  & \checkmark & \checkmark\\
    &   &   SuperFast~\citep{gao2022superfast} & Event-enhanced frame interpolation &  & \checkmark & \\
    &   &   E-SAI~\citep{yu2022learning} & Synthetic aperture imaging method &  & \checkmark & \\    
    &   &   EVSNN~\citep{zhu2022event} & Potential-assisted SNN & \checkmark & \checkmark & \\
    &   &   Spiking-Fer~\citep{barchid2023spikingfer} & Deep CSNN &  & \checkmark & \\     
    &   &   Automotive Detection~\citep{cordone2022object} &  PLIF \& SG \&  Event encoding  & \checkmark &  & \checkmark\\
    &   &   STNet~\citep{zhang2022spiking} & Spiking transformer network &  & \checkmark & \\
    &   &   LaneSNNs~\citep{viale2022lanesnns} & offline supervised learning rule &  &  & \checkmark\\  
    &   &   HALSIE~\citep{biswas2023halsie} & Hybrid approach &  & \checkmark & \\ 
    &   &   SpikeMS~\citep{parameshwara2021spikems} & Spatio-temporal loss &  &  & \checkmark\\  
    &   &   Event-based Pose Tracking~\citep{zou2023eventbased} & Spiking Spatiotemporal Transformer &  & \checkmark & \\    
\bottomrule
\end{tabular}}
%\end{tabular}
\end{center}
\end{table*}

\subsection{Efficiency Improvement Methods}

An important reason why have SNNs received extensive attention recently is that they are seen as more energy efficient than ANNs due to their event-driven computation mechanism and the replacement of energy-consuming weight multiplication with addition. To further explore the efficiency advantages of SNNs so that they can be applied to energy-constrained devices is also a hot topic in the SNN field. This kind of method can be mainly categorized into network compression techniques and sparse SNNs.
 
\subsubsection{Network compression techniques}

Network compression techniques have been widely used in ANNs. There are also some works applying these techniques in SNNs. In the literature, approaches for compressing deep SNNs can be classified into three categories: parameter pruning, NAS, and knowledge distillation. 

\textbf{Parameter pruning}. Parameter pruning mainly focuses on eliminating the redundant parameters in the model by removing the uncritical ones. SNNs, unlike their non-spiking counterparts, consist of a temporal dimension. Along with considering temporal information, a spatial and temporal pruning of SNNs is proposed in ~\citep{2021Spatio}. Generally speaking, pruning will cause accuracy degradation to some extent. To avoid this, SD-SNN~\citep{han2022adaptive} and Grad R ~\citep{chen2021pruning} proposed the pruning-regeneration method for removing the redundancy in SNNs from the brain development plasticity mechanism. With synaptic regeneration, these works can effectively prevent and repair over-pruning. Recently, an interesting temporal pruning, which is specific for SNNs, was proposed in~\citep{Chowdhury2022}. This method starts with an SNN of $T$ timesteps and reduces $T$ every iteration of training, which results in a continuum of accurate and efficient SNNs from $T$ timesteps, down to $1$ timestep. 

\textbf{Neural Architecture Searching (NAS)}. Obviously, a compact network carefully designed can reduce the storage and computation complexity of SNNs. However, due to the limitations of humans’ inherent knowledge, it is difficult for people to jump out of their original thinking paradigm and design an optimal compact model. Therefore, there are some works using NAS techniques to let the algorithm automatically design the compact neural architecture~\citep{na2022autosnn,Kim2022Neural}. Furthermore, in~\citep{kim2022exploring}, the lottery ticket hypothesis was investigated which shows that dense SNN networks contain smaller SNN subnetworks, \textit{i.e.}, winning tickets, which can achieve comparable performance to the dense ones, and the smaller compact one is picked as to be used network.

\textbf{Knowledge distillation}. The knowledge distillation methods aim at obtaining a compact model from a large model.
% (zlw question: distill a compact one from a large one, and the compact one can reproduce the large one? teacher->student->teacher)
%are designed to distill a more compact model to reproduce a larger network. 
In~\citep{kushawaha2021distilling}, a larger teacher SNN model is used to distill a smaller SNN model. And in~\citep{yang2022training,takuya2021training,xu2023constructing,xu2023biologically}, the same architecture ANN-teacher is used to distill SNN-student.

\subsubsection{Sparse SNNs}
Different from ANNs, SNNs transmit information by spike events, and the computation occurs only when the neuron receives spike events. Benefitting from this event-driven computation mechanism, SNNs can greatly save energy and run efficiently when implemented on neuromorphic hardware. Hence, limiting the firing rate of spiking neurons to achieve a sparse SNN is also a widely used way to improve the efficiency of the SNN. These kinds of methods can limit the firing rate of the SNN on both the neuron level and training technique level.

\textbf{On the neuron level.} In ASNN~\citep{zambrano2016fast}, an adaptive SNN based on a group of adaptive spiking neurons was proposed. These adaptive spiking neurons can optimize their firing rate using asynchronous pulsed Sigma-Delta coding efficiently.

\textbf{On the training technique level.} In~\citep{HanCorrelation2022}, a correlation-based regularizer, which is incorporated into a loss function, was proposed to minimize the redundancies between the features at each layer for structural sparsity. Obviously, this method is beneficial for energy-efficient.
Superspike~\citep{2017SuperSpike} added a heterosynaptic regularization term to the learning rule of the hidden layer weights to avoid pathologically high firing rates.
RecDis-SNN~\citep{Guo_2022_CVPR} incorporated a membrane potential loss into the SNN to regulate the membrane potential distribution to an appropriate range to avoid high firing rates. 
In~\citep{Pellegrini2021Low}, to enforce sparse spiking activity, a $l_1$ or $l_2$ regularization on the total number of spikes emitted by each layer was applied.

\subsection{Temporal Dynamics Utilization Methods}

Different from ANNs, SNNs enjoy rich temporal dynamics characteristics, which makes them more suitable for some particular temporal tasks and some vision sensors with high resolution in time, \textit{e.g.}, neuromorphic cameras, which can capture temporally rich information asynchronously inspired by the information process form of eyes. Given such characteristics, a great number of methods falling in sequential learning and cooperating with neuromorphic cameras have been proposed for SNNs.

\subsubsection{Sequential learning}

As aforementioned in Section~\ref{Preliminary}, SNNs maintain a dynamic state in the neuron memory. In~\citep{ponghiran2022spiking}, the usefulness of the inherent recurrence dynamics of the SNN for sequential learning was demonstrated, that it can retain important information. Thus, SNNs show better performance on sequential learning compared to ANNs with similar scales in many works. In~\citep{she2021sequence}, a function approximation theoretical basis was developed that any spike-sequence-to-spike-sequence mapping functions can be approximated by an SNN with one neuron per layer using skip-layer connections. And then, based on the basis, a suitable SNN model for the classification of spatio-temporal data was designed. In~\citep{li2022wearable}, SNNs were leveraged to study the Human Activity Recognition (HAR) task. Since SNNs allow spatio-temporal extraction of features and enjoy low-power computation with binary spikes, they can reduce up to 94\% energy consumption while achieving better accuracy compared with homogeneous ANN counterparts. In~\citep{NomuraRobustness2022}, an interesting phenomenon was found that SNNs trained with the appropriate temporal penalty settings are more robust against adversarial images than ANNs.

As the common sequential signal, many preliminary works on speech recognition systems based on spiking neural networks have been explored ~\citep{TavanaeiBio2017,TAVANAEI2017191,2018ABiologically,articleWu2020,ZhangMPD2019,2018ABiologicallyWu,JibinSpiking2018,wu2019robust}.
In~\citep{articleWu2020}, a deep spiking neural network was trained by the tandem learning method to handle the large vocabulary automatic speech recognition task. The experimental results demonstrated that the  deep SNN trained could compete with its ANN counterpart while requiring as low as 0.68 times total synaptic operations to their ANN counterparts.
There are also some works training deep SNN directly with SG methods for the speech task. 
In~\citep{ponghiran2022spiking}, inspired by the LSTM, a custom version of SNNs was defined that combines a forget gate with multi-bit outputs instead of binary spikes, yielding better accuracy than that of LSTMs, but with 2× fewer parameters.
In~\citep{bittar2023surrogate}, the spiking neural networks trained like  recurrent neural networks only using the standard surrogate gradient method can achieve promising results on speech recognition tasks, which shows the advantage of SNNs to handle this kind of task.
In~\citep{articleBittar2022}, a combination of adaptation, recurrence, and surrogate gradient techniques for spiking neural networks was proposed. And with these improvements, light spiking architectures that are not only able to compete with ANN solutions but also retain a high degree of compatibility with them were yielded. 
In~\citep{Pellegrini2021Low}, the dilated convolution spiking layers and a new regularization term to penalize the averaged number of spikes were used to train low-activity supervised convolutional spiking neural networks. The results showed that the SNN models can reach an error rate very close to standard DNNs while very energy efficient for speech tasks.
In~\cite{Sadovsky2023Speech}, a new technique for speech recognition that combines convolutional neural networks with spiking neural networks was presented to create an SNNCNN model. The results showed that the combination of CNNs and SNNs outperforms both MLPs and ANNs, providing a new route to further improvements in the field.
In~\citep{articleYin2021}, an activity-regularizing surrogate gradient method combined with recurrent networks of tunable and adaptive spiking neurons for SNNs was proposed, and the method performed well on the speech recognition task.

\subsubsection{Cooperating with neuromorphic cameras}

Neuromorphic camera, which is also called event-based cameras, have recently shown great potential for high-speed motion estimation owing to their ability to capture temporally rich information asynchronously. SNNs, with their spatio-temporal and event-driven processing mechanisms, are very suitable for handling such asynchronous data. Many excellent works combine SNNs and neuromorphic cameras to solve real-world large-scale problems. % zlw: a mount of real-world practical problems(大量的真实世界中的实际问题)?
In~\citep{hagenaars2021self,kosta2022adaptive}, an event-based optical flow estimation method was presented. In StereoSpike~\citep{ranccon2021stereospike}  a depth estimation method was provided.
SuperFast~\citep{gao2022superfast} leveraged an SNN and an event camera to present an event-enhanced high-speed video frame interpolation method. SuperFast can generate a very high frame rate (up to 5000 FPS) video from the input low frame rate (25 FPS) video. Furthermore,
Based on a hybrid network composed of SNNs and ANNs, E-SAI~\citep{yu2022learning} provided a novel synthetic aperture imaging method, which can see through dense occlusions and extreme lighting conditions from event data.
And in EVSNN~\citep{zhu2022event} a novel Event-based Video reconstruction framework was proposed. To fully use the information from different modalities, HALSIE~\citep{biswas2023halsie} proposed a hybrid approach for semantic segmentation comprised of dual encoders with an SNN branch to provide rich temporal cues from asynchronous events, and an ANN branch for extracting spatial information from regular frame data by simultaneously leveraging image and event modalities.

There are also some works that apply this technique in autonomous driving. In~\citep{cordone2022object}, fast and efficient automotive object detection with spiking neural networks on automotive event data was proposed. In \citep{zhang2022spiking}, a spiking transformer network, STNet, which can dynamically extract and fuse information from both temporal and spatial domains was proposed for single object tracking using event data. Besides, since event cameras enjoy extremely low latency and high dynamic range, they can also be used to handle the harsh environment, \textit{i.e.}, extreme lighting conditions or dense occlusions. LaneSNNs~\citep{viale2022lanesnns} presented an SNN-based approach for detecting the lanes marked on the streets using the event-based camera input. The experimental results show a  very low power consumption of about 1 W, which can significantly increase the lifetime and autonomy of battery-driven systems.

Based on the event-based cameras and SNNs, some works attempted to assist the behavioral recognition research. For examples,
Spiking-Fer~\citep{barchid2023spikingfer} proposed a new end-to-end deep convolutional SNN method to predict facial expression.
SpikeMS~\citep{parameshwara2021spikems} proposed a deep encoder-decoder SNN architecture and a novel spatio-temporal loss for motion segmentation using the event-based DVS camera as input.
In~\citep{zou2023eventbased}, a dedicated end-to-end sparse deep SNN consisting of the Spike-Element-Wise (SEW) ResNet and a novel Spiking Spatiotemporal Transformer was proposed for event-based pose tracking.  This method achieves a significant computation reduction of 80\% in FLOPS, demonstrating the superior advantage of SNN in this kind of task.

\section{Future Trends and Conclusions}\label{Conclusions}

The spiking neural networks, born in mimicking the information process of brain neurons, enjoy many specific characteristics and show great potential in many tasks, but meanwhile suffer from many weaknesses. As a consequence, a number of direct learning-based deep SNN solutions for handling these disadvantages or utilizing the advantages of SNNs have been proposed recently. As we summarized in this survey, these methods can be roughly categorized into i) accuracy improvement methods, ii) efficiency improvement methods, and iii) temporal dynamics utilization methods. Though successful milestones and progress have been achieved through these works, there are still many challenges in the field.

On the accuracy improvement aspect, the SNN still faces serious performance loss, especially for the large network and datasets. The main reasons might include:
\begin{itemize}
    \item \textit{Lack of measurement of information capacity:} it is still unclear how to precisely calculate the information capacity of the spike maps and what kind of neuron types or network topology is suitable for preserving information while the information passing through the network, even after firing function. We believe SNN neurons and architectures should not be referenced from brains or ANNs completely. Specific designs in regard to the characteristic of SNNs for preserving information should be explored. For instance, to increase the spiking neuron representative ability, the binary spike \{0, 1\}, which is used to mimic the activation or silence in the brain, can be replaced by ternary spike \{-1, 0, 1\}, thus the information capacity of the spiking neuron will be boosted, but the event-driven and multiplication-free operation advantages of the binary spike can be preserved still. And as aforementioned, the widely
used standard ResNet backbone in ANNs is not suitable for SNNs. And the PreAct ResNet backbone performs better since the membrane potential in neurons before the firing function will be added to the next block, thus the complete information will be transmitted simultaneously. While for the standard ResNet backbone, only quantized information is transmitted. To further preserve the information, adding the shortcut layer by layer in the PreAct ResNet backbone is better in our experiment, which is much different from the architectures in ANNs and is a  promising exploration direction. 
    \item \textit{Inherent optimization difficulties:} It is still a difficult problem to optimize the SNN in a discrete space, even though many novel gradient estimators or approximate functions have been proposed, there are still some huge obstacles in the field. Such as the gradient explosion/vanishing problem, with the increasing timestep, the problem along with the gradient errors will become severer and make the network hard to converge. Thus how to completely eliminate the impact of this problem to directly train an SNN with large timesteps is still under exploration.  We believe more theoretical studies and practical tricks will emerge to answer this question in the future.
\end{itemize}
It is also worth noting that accuracy is not the only criterion of SNNs, the versatility is another key criterion, that measures whether a method can be used in practice. Some methods proposed in prior works are very versatile, such as learnable spike factors proposed in Real Spike~\citep{guo2022real}, membrane potential rectifier proposed in InfLoR-SNN~\citep{Guo2022eccv}, temporal regularization loss proposed in TET~\citep{deng2022temporal}, {\it etc}. These methods enjoy simple implementation and low coupling, thus having become common widely used practices to improve the accuracy of SNNs. 
Some methods improve the accuracy of SNNs by designing complex spiking neurons or specific architectures. Such improvements usually show a stronger ability to increase performance. However, as we have
pointed out before, some of them suffer complicated computation and even lose the energy-efficiency advantage, which violates the original intention of SNNs. Therefore, purely pursuing high accuracy without considering versatility has limited significance in practice. The balance between accuracy and versatility is also an essential criterion for SNN research that should be considered in the following works.

On the efficiency improvement aspect, some prior works ignore the important fact, that the event-driven paradigm and friendly to the neuromorphic hardware make SNNs much different from ANNs. When implemented on the neuromorphic hardware, the computation in the SNN occurs only if the spiking neuron receives spike events. Hence, the direct reason for improving the efficiency of the SNN is reducing the number of the firing spikes, not reducing network size. Some methods intending to improve the efficiency of SNNs by pruning inactive neurons as doing in ANNs can not make sense in this situation. 
We even think that under the condition the SNN network size  does not exceed the capacity of the neuromorphic hardware, enlarging the network size but limiting the the number of the firing spikes at the same time may be a potential route to improve the accuracy and efficiency  simultaneously. In this way, different weights of the SNN may respond to different data, thus being equivalent to improving the representative capabilities of the SNN. However, a more systematic study needs to be done in the future.

On the temporal dynamics utilization aspect, a great number of interesting methods have been proposed and shown wide success. We think it is a very potential direction in the SNN field. Some explainable machine learning-related study indicates that different network types follow different patterns and enjoy different advantages. In this sense, it might be more meaningful to dive into the temporal dynamics of the SNN deeply, but not to pursue higher accuracy as ANNs. Meanwhile, considering the respective advantages, to use ANNs and SNNs together needs to be studied further.

Last but not least, more special applications for SNNs also should be explored still. Though SNNs have been used widely in many fields, including the neuromorphic camera, HAR task, speech recognition, autonomous driving, \textit{etc}, as aforementioned and the object detection ~\citep{2019SpikingYOLO,2020DeepSCNN}, object tracking~\citep{2020SiamSNN}, image segmentation\citep{2021ASegmentation}, robotic~\citep{Dupeyroux_2021,2020Towardsneuromorphic}, \textit{etc}, where some remarkable studies have applied SNNs on recently, compared to ANNs, their real-world applications are still very limited. considering the unique advantage, efficiency of SNNs, we think there is a great opportunity for applying SNNs in the Green Artificial Intelligence (GAI), which has become an important subfield of Artificial Intelligence and has notable practical value. We believe many studies focusing on using SNNs for GAI will emerge soon.

\section*{Conflict of Interest Statement}
%All financial, commercial or other relationships that might be perceived by the academic community as representing a potential conflict of interest must be disclosed. If no such relationship exists, authors will be asked to confirm the following statement: 

The authors declare that they have no known competing financial interests or personal relationships that could have appeared to influence the work reported in this paper.

\section*{Author Contributions}

Yufei Guo and Xuhui Huang  wrote the paper with Zhe Ma being active contributors toward editing and revising the paper as well as supervising the project. All authors contributed to the article and approved the
submitted version.

\section*{Funding}
This work is supported by grants from the National Natural Science Foundation of China under
contracts No.12202412 and No.12202413.

\bibliographystyle{Frontiers-Harvard} %  Many Frontiers journals use the Harvard referencing system (Author-date), to find the style and resources for the journal you are submitting to: https://zendesk.frontiersin.org/hc/en-us/articles/360017860337-Frontiers-Reference-Styles-by-Journal. For Humanities and Social Sciences articles please include page numbers in the in-text citations 
\bibliography{test}

\begin{thebibliography}{123}
\providecommand{\natexlab}[1]{#1}
\expandafter\ifx\csname urlstyle\endcsname\relax
  \providecommand{\doi}[1]{doi:\discretionary{}{}{}#1}\else
  \providecommand{\doi}{doi:\discretionary{}{}{}\begingroup
  \urlstyle{rm}\Url}\fi
\providecommand{\selectlanguage}[1]{\relax}
\providecommand{\bibAnnoteFile}[1]{%
  \IfFileExists{#1}{\begin{quotation}\noindent\textsc{Key:} #1\\
  \textsc{Annotation:}\ \input{#1}\end{quotation}}{}}
\providecommand{\bibAnnote}[2]{%
  \begin{quotation}\noindent\textsc{Key:} #1\\
  \textsc{Annotation:}\ #2\end{quotation}}

\bibitem[{Barchid et~al.(2023)Barchid, Allaert, Aissaoui, Mennesson, and
  Dj{\'e}raba}]{barchid2023spikingfer}
Barchid, S., Allaert, B., Aissaoui, A., Mennesson, J., and Dj{\'e}raba, C.
  (2023).
\newblock Spiking-fer: Spiking neural network for facial expression recognition
  with event cameras.
\newblock \emph{arXiv preprint arXiv:2304.10211}
\bibAnnoteFile{barchid2023spikingfer}

\bibitem[{Bellec et~al.(2018)Bellec, Salaj, Subramoney, Legenstein, and
  Maass}]{2018Long}
Bellec, G., Salaj, D., Subramoney, A., Legenstein, R., and Maass, W. (2018).
\newblock Long short-term memory and learning-to-learn in networks of spiking
  neurons.
\newblock \emph{Advances in neural information processing systems} 31
\bibAnnoteFile{2018Long}

\bibitem[{Biswas et~al.(2022)Biswas, Kosta, Liyanagedera, Apolinario, and
  Roy}]{biswas2023halsie}
Biswas, S.~D., Kosta, A., Liyanagedera, C., Apolinario, M., and Roy, K. (2022).
\newblock Halsie--hybrid approach to learning segmentation by simultaneously
  exploiting image and event modalities.
\newblock \emph{arXiv preprint arXiv:2211.10754}
\bibAnnoteFile{biswas2023halsie}

\bibitem[{Bittar and Garner(2022{\natexlab{a}})}]{articleBittar2022}
Bittar, A. and Garner, P. (2022{\natexlab{a}}).
\newblock A surrogate gradient spiking baseline for speech command recognition.
\newblock \emph{Frontiers in Neuroscience} 16, 865897.
\newblock \doi{10.3389/fnins.2022.865897}
\bibAnnoteFile{articleBittar2022}

\bibitem[{Bittar and Garner(2022{\natexlab{b}})}]{bittar2023surrogate}
Bittar, A. and Garner, P.~N. (2022{\natexlab{b}}).
\newblock Surrogate gradient spiking neural networks as encoders for large
  vocabulary continuous speech recognition.
\newblock \emph{arXiv preprint arXiv:2212.01187}
\bibAnnoteFile{bittar2023surrogate}

\bibitem[{Bohte(2011)}]{2011Error}
Bohte, S.~M. (2011).
\newblock Error-backpropagation in networks of fractionally predictive spiking
  neurons.
\newblock In \emph{International Conference on Artificial Neural Networks}
  (Springer), 60--68
\bibAnnoteFile{2011Error}

\bibitem[{Bohte et~al.(2002)Bohte, Kok, and La~Poutre}]{2002Error}
Bohte, S.~M., Kok, J.~N., and La~Poutre, H. (2002).
\newblock Error-backpropagation in temporally encoded networks of spiking
  neurons.
\newblock \emph{Neurocomputing} 48, 17--37
\bibAnnoteFile{2002Error}

\bibitem[{Booij and tat Nguyen(2005)}]{2005ABooij}
Booij, O. and tat Nguyen, H. (2005).
\newblock A gradient descent rule for spiking neurons emitting multiple spikes.
\newblock \emph{Information Processing Letters} 95, 552--558
\bibAnnoteFile{2005ABooij}

\bibitem[{Bu et~al.(2022)Bu, Ding, Yu, and Huang}]{bu2022optimized}
Bu, T., Ding, J., Yu, Z., and Huang, T. (2022).
\newblock Optimized potential initialization for low-latency spiking neural
  networks.
\newblock In \emph{Proceedings of the AAAI Conference on Artificial
  Intelligence}. vol.~36, 11--20
\bibAnnoteFile{bu2022optimized}

\bibitem[{Bu et~al.(2023)Bu, Fang, Ding, Dai, Yu, and Huang}]{bu2023optimal}
Bu, T., Fang, W., Ding, J., Dai, P., Yu, Z., and Huang, T. (2023).
\newblock Optimal ann-snn conversion for high-accuracy and ultra-low-latency
  spiking neural networks.
\newblock \emph{arXiv preprint arXiv:2303.04347}
\bibAnnoteFile{bu2023optimal}

\bibitem[{Che et~al.(2023)Che, Zhou, Ma, Fang, Chen, Shen
  et~al.}]{che2023autospikformer}
Che, K., Zhou, Z., Ma, Z., Fang, W., Chen, Y., Shen, S., et~al. (2023).
\newblock Auto-spikformer: Spikformer architecture search.
\newblock \emph{arXiv preprint arXiv:2306.00807}
\bibAnnoteFile{che2023autospikformer}

\bibitem[{Chen et~al.(2021)Chen, Yu, Fang, Huang, and Tian}]{chen2021pruning}
Chen, Y., Yu, Z., Fang, W., Huang, T., and Tian, Y. (2021).
\newblock Pruning of deep spiking neural networks through gradient rewiring.
\newblock \emph{arXiv preprint arXiv:2105.04916}
\bibAnnoteFile{chen2021pruning}

\bibitem[{Chen et~al.(2022)Chen, Zhang, Ren, and Qu}]{ChenGradual2022}
Chen, Y., Zhang, S., Ren, S., and Qu, H. (2022).
\newblock Gradual surrogate gradient learning in deep spiking neural networks.
\newblock In \emph{ICASSP 2022 - 2022 IEEE International Conference on
  Acoustics, Speech and Signal Processing (ICASSP)}. 8927--8931.
\newblock \doi{10.1109/ICASSP43922.2022.9746774}
\bibAnnoteFile{ChenGradual2022}

\bibitem[{Cheng et~al.(2020)Cheng, Hao, Xu, and Xu}]{2020LISNN}
Cheng, X., Hao, Y., Xu, J., and Xu, B. (2020).
\newblock Lisnn: Improving spiking neural networks with lateral interactions
  for robust object recognition.
\newblock In \emph{IJCAI}. 1519--1525
\bibAnnoteFile{2020LISNN}

\bibitem[{Chowdhury et~al.(2021)Chowdhury, Garg, and Roy}]{2021Spatio}
Chowdhury, S.~S., Garg, I., and Roy, K. (2021).
\newblock Spatio-temporal pruning and quantization for low-latency spiking
  neural networks.
\newblock In \emph{2021 International Joint Conference on Neural Networks
  (IJCNN)} (IEEE), 1--9
\bibAnnoteFile{2021Spatio}

\bibitem[{Chowdhury et~al.(2022)Chowdhury, Rathi, and Roy}]{Chowdhury2022}
Chowdhury, S.~S., Rathi, N., and Roy, K. (2022).
\newblock Towards ultra low latency spiking neural networks for vision and
  sequential tasks using temporal pruning.
\newblock In \emph{Computer Vision -- ECCV 2022}, eds. S.~Avidan, G.~Brostow,
  M.~Ciss{\'e}, G.~M. Farinella, and T.~Hassner (Cham: Springer Nature
  Switzerland), 709--726
\bibAnnoteFile{Chowdhury2022}

\bibitem[{Cordone et~al.(2022)Cordone, Miramond, and
  Thierion}]{cordone2022object}
Cordone, L., Miramond, B., and Thierion, P. (2022).
\newblock Object detection with spiking neural networks on automotive event
  data.
\newblock \emph{arXiv preprint arXiv:2205.04339}
\bibAnnoteFile{cordone2022object}

\bibitem[{Deng et~al.(2022)Deng, Li, Zhang, and Gu}]{deng2022temporal}
Deng, S., Li, Y., Zhang, S., and Gu, S. (2022).
\newblock Temporal efficient training of spiking neural network via gradient
  re-weighting.
\newblock \emph{arXiv preprint arXiv:2202.11946}
\bibAnnoteFile{deng2022temporal}

\bibitem[{Ding et~al.(2022)Ding, Dong, Heide, Ding, Zhou, Yin
  et~al.}]{ding2022biologically}
Ding, J., Dong, B., Heide, F., Ding, Y., Zhou, Y., Yin, B., et~al. (2022).
\newblock Biologically inspired dynamic thresholds for spiking neural networks.
\newblock In \emph{Advances in Neural Information Processing Systems}, eds.
  A.~H. Oh, A.~Agarwal, D.~Belgrave, and K.~Cho
\bibAnnoteFile{ding2022biologically}

\bibitem[{Duan et~al.(2022)Duan, Ding, Chen, Yu, and Huang}]{duan2022temporal}
Duan, C., Ding, J., Chen, S., Yu, Z., and Huang, T. (2022).
\newblock Temporal effective batch normalization in spiking neural networks.
\newblock In \emph{Advances in Neural Information Processing Systems}, eds.
  A.~H. Oh, A.~Agarwal, D.~Belgrave, and K.~Cho
\bibAnnoteFile{duan2022temporal}

\bibitem[{Dupeyroux et~al.(2021)Dupeyroux, Hagenaars, Paredes-Vall{\'e}s, and
  de~Croon}]{Dupeyroux_2021}
Dupeyroux, J., Hagenaars, J.~J., Paredes-Vall{\'e}s, F., and de~Croon, G.~C.
  (2021).
\newblock Neuromorphic control for optic-flow-based landing of mavs using the
  loihi processor.
\newblock In \emph{2021 IEEE International Conference on Robotics and
  Automation (ICRA)} (IEEE), 96--102
\bibAnnoteFile{Dupeyroux_2021}

\bibitem[{Fang et~al.(2021{\natexlab{a}})Fang, Yu, Chen, Huang, Masquelier, and
  Tian}]{2021Deep}
Fang, W., Yu, Z., Chen, Y., Huang, T., Masquelier, T., and Tian, Y.
  (2021{\natexlab{a}}).
\newblock Deep residual learning in spiking neural networks.
\newblock \emph{Advances in Neural Information Processing Systems} 34,
  21056--21069
\bibAnnoteFile{2021Deep}

\bibitem[{Fang et~al.(2021{\natexlab{b}})Fang, Yu, Chen, Masquelier, Huang, and
  Tian}]{2020Incorporating}
Fang, W., Yu, Z., Chen, Y., Masquelier, T., Huang, T., and Tian, Y.
  (2021{\natexlab{b}}).
\newblock Incorporating learnable membrane time constant to enhance learning of
  spiking neural networks.
\newblock In \emph{Proceedings of the IEEE/CVF International Conference on
  Computer Vision}. 2661--2671
\bibAnnoteFile{2020Incorporating}

\bibitem[{Feng et~al.(2022)Feng, Liu, Tang, Ma, and Pan}]{FengLTM022}
Feng, L., Liu, Q., Tang, H., Ma, D., and Pan, G. (2022).
\newblock Multi-level firing with spiking ds-resnet: Enabling better and deeper
  directly-trained spiking neural networks.
\newblock In \emph{Proceedings of the Thirty-First International Joint
  Conference on Artificial Intelligence, {IJCAI} 2022, Vienna, Austria, 23-29
  July 2022}, ed. L.~D. Raedt (ijcai.org), 2471--2477.
\newblock \doi{10.24963/ijcai.2022/343}
\bibAnnoteFile{FengLTM022}

\bibitem[{Gao et~al.(2022)Gao, Li, Li, Guo, and Dai}]{gao2022superfast}
Gao, Y., Li, S., Li, Y., Guo, Y., and Dai, Q. (2022).
\newblock Superfast: 200x video frame interpolation via event camera.
\newblock \emph{IEEE Transactions on Pattern Analysis and Machine Intelligence}
\bibAnnoteFile{gao2022superfast}

\bibitem[{Guo et~al.(2022{\natexlab{a}})Guo, Chen, Zhang, Liu, Wang, Huang
  et~al.}]{guo2022imloss}
Guo, Y., Chen, Y., Zhang, L., Liu, X., Wang, Y., Huang, X., et~al.
  (2022{\natexlab{a}}).
\newblock {IM}-loss: Information maximization loss for spiking neural networks.
\newblock In \emph{Advances in Neural Information Processing Systems}, eds.
  A.~H. Oh, A.~Agarwal, D.~Belgrave, and K.~Cho
\bibAnnoteFile{guo2022imloss}

\bibitem[{Guo et~al.(2022{\natexlab{b}})Guo, Chen, Zhang, Wang, Liu, Tong
  et~al.}]{Guo2022eccv}
Guo, Y., Chen, Y., Zhang, L., Wang, Y., Liu, X., Tong, X., et~al.
  (2022{\natexlab{b}}).
\newblock Reducing information loss for spiking neural networks.
\newblock In \emph{Computer Vision -- ECCV 2022}, eds. S.~Avidan, G.~Brostow,
  M.~Ciss{\'e}, G.~M. Farinella, and T.~Hassner (Cham: Springer Nature
  Switzerland), 36--52
\bibAnnoteFile{Guo2022eccv}

\bibitem[{Guo et~al.(2023{\natexlab{a}})Guo, Liu, Chen, Zhang, Peng, Zhang
  et~al.}]{guo2023rmploss}
Guo, Y., Liu, X., Chen, Y., Zhang, L., Peng, W., Zhang, Y., et~al.
  (2023{\natexlab{a}}).
\newblock Rmp-loss: Regularizing membrane potential distribution for spiking
  neural networks.
\newblock \emph{arXiv preprint arXiv:2308.06787}
\bibAnnoteFile{guo2023rmploss}

\bibitem[{Guo et~al.(2023{\natexlab{b}})Guo, Peng, Chen, Zhang, Liu, Huang
  et~al.}]{guo2023joint}
Guo, Y., Peng, W., Chen, Y., Zhang, L., Liu, X., Huang, X., et~al.
  (2023{\natexlab{b}}).
\newblock Joint a-snn: Joint training of artificial and spiking neural networks
  via self-distillation and weight factorization.
\newblock \emph{Pattern Recognition} , 109639
\bibAnnoteFile{guo2023joint}

\bibitem[{Guo et~al.(2022{\natexlab{c}})Guo, Tong, Chen, Zhang, Liu, Ma
  et~al.}]{Guo_2022_CVPR}
Guo, Y., Tong, X., Chen, Y., Zhang, L., Liu, X., Ma, Z., et~al.
  (2022{\natexlab{c}}).
\newblock Recdis-snn: Rectifying membrane potential distribution for directly
  training spiking neural networks.
\newblock In \emph{Proceedings of the IEEE/CVF Conference on Computer Vision
  and Pattern Recognition (CVPR)}. 326--335
\bibAnnoteFile{Guo_2022_CVPR}

\bibitem[{Guo et~al.(2022{\natexlab{d}})Guo, Zhang, Chen, Tong, Liu, Wang
  et~al.}]{guo2022real}
Guo, Y., Zhang, L., Chen, Y., Tong, X., Liu, X., Wang, Y., et~al.
  (2022{\natexlab{d}}).
\newblock Real spike: Learning real-valued spikes for spiking neural networks.
\newblock In \emph{Computer Vision--ECCV 2022: 17th European Conference, Tel
  Aviv, Israel, October 23--27, 2022, Proceedings, Part XII} (Springer), 52--68
\bibAnnoteFile{guo2022real}

\bibitem[{Guo et~al.(2023{\natexlab{c}})Guo, Zhang, Chen, Peng, Liu, Zhang
  et~al.}]{guo2023membrane}
Guo, Y., Zhang, Y., Chen, Y., Peng, W., Liu, X., Zhang, L., et~al.
  (2023{\natexlab{c}}).
\newblock Membrane potential batch normalization for spiking neural networks.
\newblock \emph{arXiv preprint arXiv:2308.08359}
\bibAnnoteFile{guo2023membrane}

\bibitem[{Hagenaars et~al.(2021)Hagenaars, Paredes-Vall{\'e}s, and
  De~Croon}]{hagenaars2021self}
Hagenaars, J., Paredes-Vall{\'e}s, F., and De~Croon, G. (2021).
\newblock Self-supervised learning of event-based optical flow with spiking
  neural networks.
\newblock \emph{Advances in Neural Information Processing Systems} 34,
  7167--7179
\bibAnnoteFile{hagenaars2021self}

\bibitem[{Han and Roy(2020)}]{2020Deep}
Han, B. and Roy, K. (2020).
\newblock Deep spiking neural network: Energy efficiency through time based
  coding.
\newblock In \emph{European Conference on Computer Vision} (Springer), 388--404
\bibAnnoteFile{2020Deep}

\bibitem[{Han et~al.(2022)Han, Zhao, Zeng, and Pan}]{han2022adaptive}
Han, B., Zhao, F., Zeng, Y., and Pan, W. (2022).
\newblock Adaptive sparse structure development with pruning and regeneration
  for spiking neural networks.
\newblock \emph{arXiv preprint arXiv:2211.12219}
\bibAnnoteFile{han2022adaptive}

\bibitem[{Han and Lee(2022)}]{HanCorrelation2022}
Han, C.~S. and Lee, K.~M. (2022).
\newblock Correlation-based regularization for fast and energy-efficient
  spiking neural networks.
\newblock In \emph{Proceedings of the 37th ACM/SIGAPP Symposium on Applied
  Computing} (New York, NY, USA: Association for Computing Machinery), SAC '22,
  1048–1055
\bibAnnoteFile{HanCorrelation2022}

\bibitem[{Hao et~al.(2020)Hao, Huang, Dong, and Xu}]{2018ABiologically}
Hao, Y., Huang, X., Dong, M., and Xu, B. (2020).
\newblock A biologically plausible supervised learning method for spiking
  neural networks using the symmetric stdp rule.
\newblock \emph{Neural Networks} 121, 387--395
\bibAnnoteFile{2018ABiologically}

\bibitem[{Hong et~al.(2019)Hong, Wei, Wang, Deng, Yu, and Che}]{2017Training}
Hong, C., Wei, X., Wang, J., Deng, B., Yu, H., and Che, Y. (2019).
\newblock Training spiking neural networks for cognitive tasks: A versatile
  framework compatible with various temporal codes.
\newblock \emph{IEEE transactions on neural networks and learning systems} 31,
  1285--1296
\bibAnnoteFile{2017Training}

\bibitem[{Hu et~al.(2021)Hu, Wu, Deng, and Li}]{2021Advancing}
Hu, Y., Wu, Y., Deng, L., and Li, G. (2021).
\newblock Advancing residual learning towards powerful deep spiking neural
  networks.
\newblock \emph{arXiv preprint arXiv:2112.08954}
\bibAnnoteFile{2021Advancing}

\bibitem[{Ikegawa et~al.(2022)Ikegawa, Saiin, Sawada, and
  Natori}]{Rethinking2022}
Ikegawa, S.-i., Saiin, R., Sawada, Y., and Natori, N. (2022).
\newblock Rethinking the role of normalization and residual blocks for spiking
  neural networks.
\newblock \emph{Sensors} 22.
\newblock \doi{10.3390/s22082876}
\bibAnnoteFile{Rethinking2022}

\bibitem[{Kim et~al.(2020)Kim, Park, Na, and Yoon}]{2019SpikingYOLO}
Kim, S., Park, S., Na, B., and Yoon, S. (2020).
\newblock Spiking-yolo: spiking neural network for energy-efficient object
  detection.
\newblock In \emph{Proceedings of the AAAI conference on artificial
  intelligence}. vol.~34, 11270--11277
\bibAnnoteFile{2019SpikingYOLO}

\bibitem[{Kim et~al.(2022{\natexlab{a}})Kim, Li, Park, Venkatesha, and
  Panda}]{Kim2022Neural}
Kim, Y., Li, Y., Park, H., Venkatesha, Y., and Panda, P. (2022{\natexlab{a}}).
\newblock Neural architecture search for spiking neural networks.
\newblock In \emph{Computer Vision--ECCV 2022: 17th European Conference, Tel
  Aviv, Israel, October 23--27, 2022, Proceedings, Part XXIV} (Springer),
  36--56
\bibAnnoteFile{Kim2022Neural}

\bibitem[{Kim et~al.(2022{\natexlab{b}})Kim, Li, Park, Venkatesha, Yin, and
  Panda}]{kim2022exploring}
Kim, Y., Li, Y., Park, H., Venkatesha, Y., Yin, R., and Panda, P.
  (2022{\natexlab{b}}).
\newblock Exploring lottery ticket hypothesis in spiking neural networks.
\newblock In \emph{European Conference on Computer Vision} (Springer), 102--120
\bibAnnoteFile{kim2022exploring}

\bibitem[{Kim and Panda(2021)}]{2020Revisiting}
Kim, Y. and Panda, P. (2021).
\newblock Revisiting batch normalization for training low-latency deep spiking
  neural networks from scratch.
\newblock \emph{Frontiers in neuroscience} , 1638
\bibAnnoteFile{2020Revisiting}

\bibitem[{Kosta and Roy(2022)}]{kosta2022adaptive}
Kosta, A.~K. and Roy, K. (2022).
\newblock Adaptive-spikenet: Event-based optical flow estimation using spiking
  neural networks with learnable neuronal dynamics.
\newblock \emph{arXiv preprint arXiv:2209.11741}
\bibAnnoteFile{kosta2022adaptive}

\bibitem[{Kushawaha et~al.(2021)Kushawaha, Kumar, Banerjee, and
  Velmurugan}]{kushawaha2021distilling}
Kushawaha, R.~K., Kumar, S., Banerjee, B., and Velmurugan, R. (2021).
\newblock Distilling spikes: Knowledge distillation in spiking neural networks.
\newblock In \emph{2020 25th International Conference on Pattern Recognition
  (ICPR)} (IEEE), 4536--4543
\bibAnnoteFile{kushawaha2021distilling}

\bibitem[{Leng et~al.(2022)Leng, Che, Zhang, Zhang, Meng, Cheng
  et~al.}]{leng2022differentiable}
Leng, L., Che, K., Zhang, K., Zhang, J., Meng, Q., Cheng, J., et~al. (2022).
\newblock Differentiable hierarchical and surrogate gradient search for spiking
  neural networks.
\newblock In \emph{Advances in Neural Information Processing Systems}, eds.
  A.~H. Oh, A.~Agarwal, D.~Belgrave, and K.~Cho
\bibAnnoteFile{leng2022differentiable}

\bibitem[{Li et~al.(2022{\natexlab{a}})Li, Chen, Guo, Zhang, and
  Wang}]{2022Brain}
Li, W., Chen, H., Guo, J., Zhang, Z., and Wang, Y. (2022{\natexlab{a}}).
\newblock Brain-inspired multilayer perceptron with spiking neurons.
\newblock In \emph{Proceedings of the IEEE/CVF Conference on Computer Vision
  and Pattern Recognition}. 783--793
\bibAnnoteFile{2022Brain}

\bibitem[{Li et~al.(2021{\natexlab{a}})Li, Deng, Dong, Gong, and
  Gu}]{li2021free}
Li, Y., Deng, S., Dong, X., Gong, R., and Gu, S. (2021{\natexlab{a}}).
\newblock A free lunch from ann: Towards efficient, accurate spiking neural
  networks calibration.
\newblock In \emph{International Conference on Machine Learning} (PMLR),
  6316--6325
\bibAnnoteFile{li2021free}

\bibitem[{Li et~al.(2021{\natexlab{b}})Li, Guo, Zhang, Deng, Hai, and
  Gu}]{li2021differentiable}
Li, Y., Guo, Y., Zhang, S., Deng, S., Hai, Y., and Gu, S. (2021{\natexlab{b}}).
\newblock Differentiable spike: Rethinking gradient-descent for training
  spiking neural networks.
\newblock \emph{Advances in Neural Information Processing Systems} 34,
  23426--23439
\bibAnnoteFile{li2021differentiable}

\bibitem[{Li et~al.(2022{\natexlab{b}})Li, Yin, Park, Kim, and
  Panda}]{li2022wearable}
Li, Y., Yin, R., Park, H., Kim, Y., and Panda, P. (2022{\natexlab{b}}).
\newblock Wearable-based human activity recognition with spatio-temporal
  spiking neural networks.
\newblock \emph{arXiv preprint arXiv:2212.02233}
\bibAnnoteFile{li2022wearable}

\bibitem[{Li and Zeng(2022)}]{li2022efficient}
Li, Y. and Zeng, Y. (2022).
\newblock Efficient and accurate conversion of spiking neural network with
  burst spikes.
\newblock \emph{arXiv preprint arXiv:2204.13271}
\bibAnnoteFile{li2022efficient}

\bibitem[{Liu et~al.(2022)Liu, Zhao, Chen, Wang, and
  Jiang}]{liu2022spikeconverter}
Liu, F., Zhao, W., Chen, Y., Wang, Z., and Jiang, L. (2022).
\newblock Spikeconverter: An efficient conversion framework zipping the gap
  between artificial neural networks and spiking neural networks.
\newblock In \emph{Proceedings of the AAAI Conference on Artificial
  Intelligence}. vol.~36, 1692--1701
\bibAnnoteFile{liu2022spikeconverter}

\bibitem[{Lobov et~al.(2020)Lobov, Mikhaylov, Shamshin, Makarov, and
  Kazantsev}]{2020Spatial}
Lobov, S.~A., Mikhaylov, A.~N., Shamshin, M., Makarov, V.~A., and Kazantsev,
  V.~B. (2020).
\newblock Spatial properties of stdp in a self-learning spiking neural network
  enable controlling a mobile robot.
\newblock \emph{Frontiers in neuroscience} 14, 88
\bibAnnoteFile{2020Spatial}

\bibitem[{Luo et~al.(2022)Luo, Qu, Wang, Yi, Zhang, and
  Zhang}]{SupervisedLuo2022}
Luo, X., Qu, H., Wang, Y., Yi, Z., Zhang, J., and Zhang, M. (2022).
\newblock Supervised learning in multilayer spiking neural networks with spike
  temporal error backpropagation.
\newblock \emph{IEEE Transactions on Neural Networks and Learning Systems} ,
  1--13\doi{10.1109/TNNLS.2022.3164930}
\bibAnnoteFile{SupervisedLuo2022}

\bibitem[{Luo et~al.(2020)Luo, Xu, Yuan, Cao, Xu, Wang et~al.}]{2020SiamSNN}
Luo, Y., Xu, M., Yuan, C., Cao, X., Xu, Y., Wang, T., et~al. (2020).
\newblock Siamsnn: Spike-based siamese network for energy-efficient and
  real-time object tracking.
\newblock \emph{arXiv preprint arXiv:2003.07584}
\bibAnnoteFile{2020SiamSNN}

\bibitem[{Ma et~al.(2023)Ma, Yan, and Tang}]{ma2023exploiting}
Ma, G., Yan, R., and Tang, H. (2023).
\newblock Exploiting noise as a resource for computation and learning in
  spiking neural networks.
\newblock \emph{arXiv preprint arXiv:2305.16044}
\bibAnnoteFile{ma2023exploiting}

\bibitem[{Meng et~al.(2022)Meng, Xiao, Yan, Wang, Lin, and
  Luo}]{meng2022training}
Meng, Q., Xiao, M., Yan, S., Wang, Y., Lin, Z., and Luo, Z.-Q. (2022).
\newblock Training high-performance low-latency spiking neural networks by
  differentiation on spike representation.
\newblock In \emph{Proceedings of the IEEE/CVF Conference on Computer Vision
  and Pattern Recognition}. 12444--12453
\bibAnnoteFile{meng2022training}

\bibitem[{Na et~al.(2022)Na, Mok, Park, Lee, Choe, and Yoon}]{na2022autosnn}
Na, B., Mok, J., Park, S., Lee, D., Choe, H., and Yoon, S. (2022).
\newblock Autosnn: Towards energy-efficient spiking neural networks.
\newblock \emph{arXiv preprint arXiv:2201.12738}
\bibAnnoteFile{na2022autosnn}

\bibitem[{Neftci et~al.(2019)Neftci, Mostafa, and Zenke}]{2019Surrogate}
Neftci, E.~O., Mostafa, H., and Zenke, F. (2019).
\newblock Surrogate gradient learning in spiking neural networks: Bringing the
  power of gradient-based optimization to spiking neural networks.
\newblock \emph{IEEE Signal Processing Magazine} 36, 51--63
\bibAnnoteFile{2019Surrogate}

\bibitem[{Nomura et~al.(2022)Nomura, Sakemi, Hosomi, and
  Morie}]{NomuraRobustness2022}
Nomura, O., Sakemi, Y., Hosomi, T., and Morie, T. (2022).
\newblock Robustness of spiking neural networks based on time-to-first-spike
  encoding against adversarial attacks.
\newblock \emph{IEEE Transactions on Circuits and Systems II: Express Briefs}
  69, 3640--3644.
\newblock \doi{10.1109/TCSII.2022.3184313}
\bibAnnoteFile{NomuraRobustness2022}

\bibitem[{Parameshwara et~al.(2021)Parameshwara, Li, Ferm{\"u}ller, Sanket,
  Evanusa, and Aloimonos}]{parameshwara2021spikems}
Parameshwara, C.~M., Li, S., Ferm{\"u}ller, C., Sanket, N.~J., Evanusa, M.~S.,
  and Aloimonos, Y. (2021).
\newblock Spikems: Deep spiking neural network for motion segmentation.
\newblock In \emph{2021 IEEE/RSJ International Conference on Intelligent Robots
  and Systems (IROS)} (IEEE), 3414--3420
\bibAnnoteFile{parameshwara2021spikems}

\bibitem[{Patel et~al.(2021)Patel, Hunsberger, Batir, and
  Eliasmith}]{2021ASegmentation}
Patel, K., Hunsberger, E., Batir, S., and Eliasmith, C. (2021).
\newblock A spiking neural network for image segmentation.
\newblock \emph{arXiv preprint arXiv:2106.08921}
\bibAnnoteFile{2021ASegmentation}

\bibitem[{Pellegrini et~al.(2021)Pellegrini, Zimmer, and
  Masquelier}]{Pellegrini2021Low}
Pellegrini, T., Zimmer, R., and Masquelier, T. (2021).
\newblock Low-activity supervised convolutional spiking neural networks applied
  to speech commands recognition.
\newblock In \emph{2021 IEEE Spoken Language Technology Workshop (SLT)}.
  97--103.
\newblock \doi{10.1109/SLT48900.2021.9383587}
\bibAnnoteFile{Pellegrini2021Low}

\bibitem[{Ponghiran and Roy(2022)}]{ponghiran2022spiking}
Ponghiran, W. and Roy, K. (2022).
\newblock Spiking neural networks with improved inherent recurrence dynamics
  for sequential learning.
\newblock In \emph{Proceedings of the AAAI Conference on Artificial
  Intelligence}. vol.~36, 8001--8008
\bibAnnoteFile{ponghiran2022spiking}

\bibitem[{Ponulak and Kasinski(2011)}]{ponulak2011introduction}
Ponulak, F. and Kasinski, A. (2011).
\newblock Introduction to spiking neural networks: Information processing,
  learning and applications.
\newblock \emph{Acta neurobiologiae experimentalis} 71, 409--433
\bibAnnoteFile{ponulak2011introduction}

\bibitem[{Ran{\c{c}}on et~al.(2021)Ran{\c{c}}on, Cuadrado-Anibarro, Cottereau,
  and Masquelier}]{ranccon2021stereospike}
Ran{\c{c}}on, U., Cuadrado-Anibarro, J., Cottereau, B.~R., and Masquelier, T.
  (2021).
\newblock Stereospike: Depth learning with a spiking neural network.
\newblock \emph{arXiv preprint arXiv:2109.13751}
\bibAnnoteFile{ranccon2021stereospike}

\bibitem[{Rathi and Roy(2020)}]{2020DIET}
Rathi, N. and Roy, K. (2020).
\newblock Diet-snn: Direct input encoding with leakage and threshold
  optimization in deep spiking neural networks.
\newblock \emph{arXiv preprint arXiv:2008.03658}
\bibAnnoteFile{2020DIET}

\bibitem[{Roy et~al.(2019)Roy, Jaiswal, and Panda}]{roy2019towards}
Roy, K., Jaiswal, A., and Panda, P. (2019).
\newblock Towards spike-based machine intelligence with neuromorphic computing.
\newblock \emph{Nature} 575, 607--617
\bibAnnoteFile{roy2019towards}

\bibitem[{Sadovsky et~al.(2023)Sadovsky, Jakubec, and
  Jarina}]{Sadovsky2023Speech}
Sadovsky, E., Jakubec, M., and Jarina, R. (2023).
\newblock Speech command recognition based on convolutional spiking neural
  networks.
\newblock In \emph{2023 33rd International Conference Radioelektronika
  (RADIOELEKTRONIKA)}. 1--5.
\newblock \doi{10.1109/RADIOELEKTRONIKA57919.2023.10109082}
\bibAnnoteFile{Sadovsky2023Speech}

\bibitem[{She et~al.(2021)She, Dash, and Mukhopadhyay}]{she2021sequence}
She, X., Dash, S., and Mukhopadhyay, S. (2021).
\newblock Sequence approximation using feedforward spiking neural network for
  spatiotemporal learning: Theory and optimization methods.
\newblock In \emph{International Conference on Learning Representations}
\bibAnnoteFile{she2021sequence}

\bibitem[{Shen et~al.(2023)Shen, Zhao, and Zeng}]{shen2023exploiting}
Shen, G., Zhao, D., and Zeng, Y. (2023).
\newblock Exploiting high performance spiking neural networks with efficient
  spiking patterns.
\newblock \emph{arXiv preprint arXiv:2301.12356}
\bibAnnoteFile{shen2023exploiting}

\bibitem[{Stagsted et~al.(2020)Stagsted, Vitale, Binz, Renner, and
  Sandamirskaya}]{2020Towardsneuromorphic}
Stagsted, R., Vitale, A., Binz, J., Renner, A., and Sandamirskaya, Y. (2020).
\newblock Towards neuromorphic control: A spiking neural network based pid
  controller for uav.
\newblock In \emph{Robotics: Science and Systems 2020}
\bibAnnoteFile{2020Towardsneuromorphic}

\bibitem[{Takuya et~al.(2021)Takuya, Zhang, and Nakashima}]{takuya2021training}
Takuya, S., Zhang, R., and Nakashima, Y. (2021).
\newblock Training low-latency spiking neural network through knowledge
  distillation.
\newblock In \emph{2021 IEEE Symposium in Low-Power and High-Speed Chips (COOL
  CHIPS)} (IEEE), 1--3
\bibAnnoteFile{takuya2021training}

\bibitem[{Tavanaei et~al.(2019)Tavanaei, Ghodrati, Kheradpisheh, Masquelier,
  and Maida}]{tavanaei2019deep}
Tavanaei, A., Ghodrati, M., Kheradpisheh, S.~R., Masquelier, T., and Maida, A.
  (2019).
\newblock Deep learning in spiking neural networks.
\newblock \emph{Neural networks} 111, 47--63
\bibAnnoteFile{tavanaei2019deep}

\bibitem[{Tavanaei and Maida(2017{\natexlab{a}})}]{TavanaeiBio2017}
Tavanaei, A. and Maida, A. (2017{\natexlab{a}}).
\newblock Bio-inspired multi-layer spiking neural network extracts
  discriminative features from speech signals.
\newblock In \emph{Neural Information Processing}, eds. D.~Liu, S.~Xie, Y.~Li,
  D.~Zhao, and E.-S.~M. El-Alfy (Cham: Springer International Publishing),
  899--908
\bibAnnoteFile{TavanaeiBio2017}

\bibitem[{Tavanaei and Maida(2017{\natexlab{b}})}]{TAVANAEI2017191}
Tavanaei, A. and Maida, A.~S. (2017{\natexlab{b}}).
\newblock A spiking network that learns to extract spike signatures from speech
  signals.
\newblock \emph{Neurocomputing} 240, 191--199.
\newblock \doi{https://doi.org/10.1016/j.neucom.2017.01.088}
\bibAnnoteFile{TAVANAEI2017191}

\bibitem[{Viale et~al.(2022)Viale, Marchisio, Martina, Masera, and
  Shafique}]{viale2022lanesnns}
Viale, A., Marchisio, A., Martina, M., Masera, G., and Shafique, M. (2022).
\newblock Lanesnns: Spiking neural networks for lane detection on the loihi
  neuromorphic processor.
\newblock In \emph{2022 IEEE/RSJ International Conference on Intelligent Robots
  and Systems (IROS)} (IEEE), 79--86
\bibAnnoteFile{viale2022lanesnns}

\bibitem[{Wang et~al.(2022{\natexlab{a}})Wang, Cheng, and Lim}]{wang2022ltmd}
Wang, S., Cheng, T.~H., and Lim, M.-H. (2022{\natexlab{a}}).
\newblock {LTMD}: Learning improvement of spiking neural networks with
  learnable thresholding neurons and moderate dropout.
\newblock In \emph{Advances in Neural Information Processing Systems}, eds.
  A.~H. Oh, A.~Agarwal, D.~Belgrave, and K.~Cho
\bibAnnoteFile{wang2022ltmd}

\bibitem[{Wang et~al.(2020)Wang, Lin, and Dang}]{wang2020supervised}
Wang, X., Lin, X., and Dang, X. (2020).
\newblock Supervised learning in spiking neural networks: A review of
  algorithms and evaluations.
\newblock \emph{Neural Networks} 125, 258--280
\bibAnnoteFile{wang2020supervised}

\bibitem[{Wang et~al.(2023)Wang, Zhang, and Zhang}]{wang2023mtsnn}
Wang, X., Zhang, Y., and Zhang, Y. (2023).
\newblock Mt-snn: Enhance spiking neural network with multiple thresholds.
\newblock \emph{arXiv preprint arXiv:2303.11127}
\bibAnnoteFile{wang2023mtsnn}

\bibitem[{Wang et~al.(2022{\natexlab{b}})Wang, Zhang, Chen, and
  Qu}]{wang2022signed}
Wang, Y., Zhang, M., Chen, Y., and Qu, H. (2022{\natexlab{b}}).
\newblock Signed neuron with memory: Towards simple, accurate and
  high-efficient ann-snn conversion.
\newblock In \emph{International Joint Conference on Artificial Intelligence}
\bibAnnoteFile{wang2022signed}

\bibitem[{Wu et~al.(2018{\natexlab{a}})Wu, Chua, and Li}]{2018ABiologicallyWu}
Wu, J., Chua, Y., and Li, H. (2018{\natexlab{a}}).
\newblock A biologically plausible speech recognition framework based on
  spiking neural networks.
\newblock In \emph{2018 International Joint Conference on Neural Networks
  (IJCNN)}
\bibAnnoteFile{2018ABiologicallyWu}

\bibitem[{Wu et~al.(2021{\natexlab{a}})Wu, Chua, Zhang, Li, Li, and
  Tan}]{2021ATandem}
Wu, J., Chua, Y., Zhang, M., Li, G., Li, H., and Tan, K.~C.
  (2021{\natexlab{a}}).
\newblock A tandem learning rule for effective training and rapid inference of
  deep spiking neural networks.
\newblock \emph{IEEE Transactions on Neural Networks and Learning Systems}
\bibAnnoteFile{2021ATandem}

\bibitem[{Wu et~al.(2018{\natexlab{b}})Wu, Chua, Zhang, Li, and
  Tan}]{JibinSpiking2018}
Wu, J., Chua, Y., Zhang, M., Li, H., and Tan, K.~C. (2018{\natexlab{b}}).
\newblock A spiking neural network framework for robust sound classification.
\newblock \emph{Frontiers in Neuroscience} 12.
\newblock \doi{10.3389/fnins.2018.00836}
\bibAnnoteFile{JibinSpiking2018}

\bibitem[{Wu et~al.(2019{\natexlab{a}})Wu, Chua, Zhang, Yang, Li, and
  Li}]{wu2019deep}
Wu, J., Chua, Y., Zhang, M., Yang, Q., Li, G., and Li, H. (2019{\natexlab{a}}).
\newblock Deep spiking neural network with spike count based learning rule.
\newblock In \emph{2019 International Joint Conference on Neural Networks
  (IJCNN)} (IEEE), 1--6
\bibAnnoteFile{wu2019deep}

\bibitem[{Wu et~al.(2019{\natexlab{b}})Wu, Pan, Zhang, Das, Chua, and
  Li}]{wu2019robust}
Wu, J., Pan, Z., Zhang, M., Das, R.~K., Chua, Y., and Li, H.
  (2019{\natexlab{b}}).
\newblock Robust sound recognition: A neuromorphic approach.
\newblock In \emph{Interspeech}. 3667--3668
\bibAnnoteFile{wu2019robust}

\bibitem[{Wu et~al.(2021{\natexlab{b}})Wu, Xu, Han, Zhou, Zhang, Li
  et~al.}]{2020Progressive}
Wu, J., Xu, C., Han, X., Zhou, D., Zhang, M., Li, H., et~al.
  (2021{\natexlab{b}}).
\newblock Progressive tandem learning for pattern recognition with deep spiking
  neural networks.
\newblock \emph{IEEE Transactions on Pattern Analysis and Machine Intelligence}
  44, 7824--7840
\bibAnnoteFile{2020Progressive}

\bibitem[{Wu et~al.(2020)Wu, Yılmaz, Zhang, Li, and Tan}]{articleWu2020}
Wu, J., Yılmaz, E., Zhang, M., Li, H., and Tan, K. (2020).
\newblock Deep spiking neural networks for large vocabulary automatic speech
  recognition.
\newblock \emph{Frontiers in Neuroscience} 14.
\newblock \doi{10.3389/fnins.2020.00199}
\bibAnnoteFile{articleWu2020}

\bibitem[{Wu et~al.(2018{\natexlab{c}})Wu, Deng, Li, Zhu, and Shi}]{2018Spatio}
Wu, Y., Deng, L., Li, G., Zhu, J., and Shi, L. (2018{\natexlab{c}}).
\newblock Spatio-temporal backpropagation for training high-performance spiking
  neural networks.
\newblock \emph{Frontiers in neuroscience} 12, 331
\bibAnnoteFile{2018Spatio}

\bibitem[{Wu et~al.(2019{\natexlab{c}})Wu, Deng, Li, Zhu, Xie, and
  Shi}]{2018Direct}
Wu, Y., Deng, L., Li, G., Zhu, J., Xie, Y., and Shi, L. (2019{\natexlab{c}}).
\newblock Direct training for spiking neural networks: Faster, larger, better.
\newblock In \emph{Proceedings of the AAAI Conference on Artificial
  Intelligence}. vol.~33, 1311--1318
\bibAnnoteFile{2018Direct}

\bibitem[{Xu et~al.(2023{\natexlab{a}})Xu, Li, Fang, Shen, Liu, Tang
  et~al.}]{xu2023biologically}
Xu, Q., Li, Y., Fang, X., Shen, J., Liu, J.~K., Tang, H., et~al.
  (2023{\natexlab{a}}).
\newblock Biologically inspired structure learning with reverse knowledge
  distillation for spiking neural networks.
\newblock \emph{arXiv preprint arXiv:2304.09500}
\bibAnnoteFile{xu2023biologically}

\bibitem[{Xu et~al.(2023{\natexlab{b}})Xu, Li, Shen, Liu, Tang, and
  Pan}]{xu2023constructing}
Xu, Q., Li, Y., Shen, J., Liu, J.~K., Tang, H., and Pan, G.
  (2023{\natexlab{b}}).
\newblock Constructing deep spiking neural networks from artificial neural
  networks with knowledge distillation.
\newblock \emph{arXiv preprint arXiv:2304.05627}
\bibAnnoteFile{xu2023constructing}

\bibitem[{Xu et~al.(2013)Xu, Zeng, Han, and Yang}]{2013AXusupervised}
Xu, Y., Zeng, X., Han, L., and Yang, J. (2013).
\newblock A supervised multi-spike learning algorithm based on gradient descent
  for spiking neural networks.
\newblock \emph{Neural Networks} 43, 99--113
\bibAnnoteFile{2013AXusupervised}

\bibitem[{Yamazaki et~al.(2022)Yamazaki, Vo-Ho, Bulsara, and
  Le}]{yamazaki2022spiking}
Yamazaki, K., Vo-Ho, V.-K., Bulsara, D., and Le, N. (2022).
\newblock Spiking neural networks and their applications: A review.
\newblock \emph{Brain Sciences} 12, 863
\bibAnnoteFile{yamazaki2022spiking}

\bibitem[{Yang et~al.(2022)Yang, Wu, Zhang, Chua, Wang, and
  Li}]{yang2022training}
Yang, Q., Wu, J., Zhang, M., Chua, Y., Wang, X., and Li, H. (2022).
\newblock Training spiking neural networks with local tandem learning.
\newblock \emph{arXiv preprint arXiv:2210.04532}
\bibAnnoteFile{yang2022training}

\bibitem[{Yao et~al.(2021)Yao, Gao, Zhao, Wang, Lin, Yang
  et~al.}]{yao2021temporal}
Yao, M., Gao, H., Zhao, G., Wang, D., Lin, Y., Yang, Z., et~al. (2021).
\newblock Temporal-wise attention spiking neural networks for event streams
  classification.
\newblock In \emph{Proceedings of the IEEE/CVF International Conference on
  Computer Vision}. 10221--10230
\bibAnnoteFile{yao2021temporal}

\bibitem[{Yao et~al.(2022)Yao, Li, Mo, and Cheng}]{GLIF}
Yao, X., Li, F., Mo, Z., and Cheng, J. (2022).
\newblock {GLIF:} {A} unified gated leaky integrate-and-fire neuron for spiking
  neural networks.
\newblock \emph{CoRR} abs/2210.13768.
\newblock \doi{10.48550/arXiv.2210.13768}
\bibAnnoteFile{GLIF}

\bibitem[{Yin et~al.(2020)Yin, Corradi, and Boht{\'e}}]{2020Effective}
Yin, B., Corradi, F., and Boht{\'e}, S.~M. (2020).
\newblock Effective and efficient computation with multiple-timescale spiking
  recurrent neural networks.
\newblock In \emph{International Conference on Neuromorphic Systems 2020}. 1--8
\bibAnnoteFile{2020Effective}

\bibitem[{Yin et~al.(2021)Yin, Corradi, and Bohté}]{articleYin2021}
Yin, B., Corradi, F., and Bohté, S. (2021).
\newblock Accurate and efficient time-domain classification with adaptive
  spiking recurrent neural networks.
\newblock \emph{Nature Machine Intelligence} 3, 905--913.
\newblock \doi{10.1038/s42256-021-00397-w}
\bibAnnoteFile{articleYin2021}

\bibitem[{Yu et~al.(2022{\natexlab{a}})Yu, Gu, Li, Wang, Wang, and
  Li}]{yu2022stsc}
Yu, C., Gu, Z., Li, D., Wang, G., Wang, A., and Li, E. (2022{\natexlab{a}}).
\newblock Stsc-snn: Spatio-temporal synaptic connection with temporal
  convolution and attention for spiking neural networks.
\newblock \emph{arXiv preprint arXiv:2210.05241}
\bibAnnoteFile{yu2022stsc}

\bibitem[{Yu et~al.(2022{\natexlab{b}})Yu, Zhang, Liao, Yang, and
  Xia}]{yu2022learning}
Yu, L., Zhang, X., Liao, W., Yang, W., and Xia, G.-S. (2022{\natexlab{b}}).
\newblock Learning to see through with events.
\newblock \emph{IEEE Transactions on Pattern Analysis and Machine Intelligence}
\bibAnnoteFile{yu2022learning}

\bibitem[{Yu et~al.(2022{\natexlab{c}})Yu, Gao, Wei, Li, Tan, and
  Huang}]{ImprovingYu2022}
Yu, Q., Gao, J., Wei, J., Li, J., Tan, K.~C., and Huang, T.
  (2022{\natexlab{c}}).
\newblock Improving multispike learning with plastic synaptic delays.
\newblock \emph{IEEE Transactions on Neural Networks and Learning Systems} ,
  1--12\doi{10.1109/TNNLS.2022.3165527}
\bibAnnoteFile{ImprovingYu2022}

\bibitem[{Yu et~al.(2022{\natexlab{d}})Yu, Song, Ma, Pan, and
  Tan}]{SynapticYu2022}
Yu, Q., Song, S., Ma, C., Pan, L., and Tan, K.~C. (2022{\natexlab{d}}).
\newblock Synaptic learning with augmented spikes.
\newblock \emph{IEEE Transactions on Neural Networks and Learning Systems} 33,
  1134--1146.
\newblock \doi{10.1109/TNNLS.2020.3040969}
\bibAnnoteFile{SynapticYu2022}

\bibitem[{Zambrano and Bohte(2016)}]{zambrano2016fast}
Zambrano, D. and Bohte, S.~M. (2016).
\newblock Fast and efficient asynchronous neural computation with adapting
  spiking neural networks.
\newblock \emph{arXiv preprint arXiv:1609.02053}
\bibAnnoteFile{zambrano2016fast}

\bibitem[{Zenke and Ganguli(2018)}]{2017SuperSpike}
Zenke, F. and Ganguli, S. (2018).
\newblock Superspike: Supervised learning in multilayer spiking neural
  networks.
\newblock \emph{Neural computation} 30, 1514--1541
\bibAnnoteFile{2017SuperSpike}

\bibitem[{Zhang et~al.(2022{\natexlab{a}})Zhang, Zhang, Jia, Wang, and
  Xu}]{zhang2022recent}
Zhang, D., Zhang, T., Jia, S., Wang, Q., and Xu, B. (2022{\natexlab{a}}).
\newblock Recent advances and new frontiers in spiking neural networks.
\newblock \emph{arXiv preprint arXiv:2204.07050}
\bibAnnoteFile{zhang2022recent}

\bibitem[{Zhang et~al.(2022{\natexlab{b}})Zhang, Dong, Zhang, Ding, Heide, Yin
  et~al.}]{zhang2022spiking}
Zhang, J., Dong, B., Zhang, H., Ding, J., Heide, F., Yin, B., et~al.
  (2022{\natexlab{b}}).
\newblock Spiking transformers for event-based single object tracking.
\newblock In \emph{Proceedings of the IEEE/CVF Conference on Computer Vision
  and Pattern Recognition}. 8801--8810
\bibAnnoteFile{zhang2022spiking}

\bibitem[{Zhang et~al.(2020{\natexlab{a}})Zhang, Luo, Chen, Wu, Belatreche, Pan
  et~al.}]{zhang2020efficient}
Zhang, M., Luo, X., Chen, Y., Wu, J., Belatreche, A., Pan, Z., et~al.
  (2020{\natexlab{a}}).
\newblock An efficient threshold-driven aggregate-label learning algorithm for
  multimodal information processing.
\newblock \emph{IEEE Journal of Selected Topics in Signal Processing} 14,
  592--602
\bibAnnoteFile{zhang2020efficient}

\bibitem[{Zhang et~al.(2022{\natexlab{c}})Zhang, Wang, Wu, Belatreche,
  Amornpaisannon, Zhang et~al.}]{ZhangRectified2022}
Zhang, M., Wang, J., Wu, J., Belatreche, A., Amornpaisannon, B., Zhang, Z.,
  et~al. (2022{\natexlab{c}}).
\newblock Rectified linear postsynaptic potential function for backpropagation
  in deep spiking neural networks.
\newblock \emph{IEEE Transactions on Neural Networks and Learning Systems} 33,
  1947--1958.
\newblock \doi{10.1109/TNNLS.2021.3110991}
\bibAnnoteFile{ZhangRectified2022}

\bibitem[{Zhang et~al.(2020{\natexlab{b}})Zhang, Wu, Belatreche, Pan, Xie, Chua
  et~al.}]{zhang2020supervised}
Zhang, M., Wu, J., Belatreche, A., Pan, Z., Xie, X., Chua, Y., et~al.
  (2020{\natexlab{b}}).
\newblock Supervised learning in spiking neural networks with synaptic
  delay-weight plasticity.
\newblock \emph{Neurocomputing} 409, 103--118
\bibAnnoteFile{zhang2020supervised}

\bibitem[{Zhang et~al.(2019)Zhang, Wu, Chua, Luo, Pan, Liu
  et~al.}]{ZhangMPD2019}
Zhang, M., Wu, J., Chua, Y., Luo, X., Pan, Z., Liu, D., et~al. (2019).
\newblock Mpd-al: An efficient membrane potential driven aggregate-label
  learning algorithm for spiking neurons.
\newblock \emph{Proceedings of the AAAI Conference on Artificial Intelligence}
  33, 1327--1334.
\newblock \doi{10.1609/aaai.v33i01.33011327}
\bibAnnoteFile{ZhangMPD2019}

\bibitem[{Zhang and Li(2020)}]{zhang2021temporal}
Zhang, W. and Li, P. (2020).
\newblock Temporal spike sequence learning via backpropagation for deep spiking
  neural networks.
\newblock \emph{Advances in Neural Information Processing Systems} 33,
  12022--12033
\bibAnnoteFile{zhang2021temporal}

\bibitem[{Zheng et~al.(2021)Zheng, Wu, Deng, Hu, and Li}]{2020Going}
Zheng, H., Wu, Y., Deng, L., Hu, Y., and Li, G. (2021).
\newblock Going deeper with directly-trained larger spiking neural networks.
\newblock In \emph{Proceedings of the AAAI Conference on Artificial
  Intelligence}. vol.~35, 11062--11070
\bibAnnoteFile{2020Going}

\bibitem[{Zhou et~al.(2023)Zhou, Yu, Zhou, Ma, Zhang, Zhou
  et~al.}]{zhou2023spikingformer}
Zhou, C., Yu, L., Zhou, Z., Ma, Z., Zhang, H., Zhou, H., et~al. (2023).
\newblock Spikingformer: Spike-driven residual learning for transformer-based
  spiking neural network.
\newblock \emph{arXiv preprint arXiv:2304.11954}
\bibAnnoteFile{zhou2023spikingformer}

\bibitem[{Zhou et~al.(2020)Zhou, Chen, Li, and Sanyal}]{2020DeepSCNN}
Zhou, S., Chen, Y., Li, X., and Sanyal, A. (2020).
\newblock Deep scnn-based real-time object detection for self-driving vehicles
  using lidar temporal data.
\newblock \emph{IEEE Access} 8, 76903--76912
\bibAnnoteFile{2020DeepSCNN}

\bibitem[{Zhou et~al.(2021)Zhou, Li, Chen, Chandrasekaran, and
  Sanyal}]{zhou2021temporalcoded}
Zhou, S., Li, X., Chen, Y., Chandrasekaran, S.~T., and Sanyal, A. (2021).
\newblock Temporal-coded deep spiking neural network with easy training and
  robust performance.
\newblock In \emph{Proceedings of the AAAI Conference on Artificial
  Intelligence}. vol.~35, 11143--11151
\bibAnnoteFile{zhou2021temporalcoded}

\bibitem[{Zhou et~al.(2022)Zhou, Zhu, He, Wang, Yan, Tian
  et~al.}]{zhou2022spikformer}
Zhou, Z., Zhu, Y., He, C., Wang, Y., Yan, S., Tian, Y., et~al. (2022).
\newblock Spikformer: When spiking neural network meets transformer.
\newblock \emph{arXiv preprint arXiv:2209.15425}
\bibAnnoteFile{zhou2022spikformer}

\bibitem[{Zhu et~al.(2022{\natexlab{a}})Zhu, Wang, Chang, Li, Huang, and
  Tian}]{zhu2022event}
Zhu, L., Wang, X., Chang, Y., Li, J., Huang, T., and Tian, Y.
  (2022{\natexlab{a}}).
\newblock Event-based video reconstruction via potential-assisted spiking
  neural network.
\newblock In \emph{Proceedings of the IEEE/CVF Conference on Computer Vision
  and Pattern Recognition}. 3594--3604
\bibAnnoteFile{zhu2022event}

\bibitem[{Zhu et~al.(2022{\natexlab{b}})Zhu, Zhao, Zhang, Deng, Duan, Zhang
  et~al.}]{zhu2022tcjasnn}
Zhu, R.-J., Zhao, Q., Zhang, T., Deng, H., Duan, Y., Zhang, M., et~al.
  (2022{\natexlab{b}}).
\newblock Tcja-snn: Temporal-channel joint attention for spiking neural
  networks.
\newblock \emph{arXiv preprint arXiv:2206.10177}
\bibAnnoteFile{zhu2022tcjasnn}

\bibitem[{Zhu et~al.(2022{\natexlab{c}})Zhu, Yu, Fang, Xie, Huang, and
  Masquelier}]{zhu2022training}
Zhu, Y., Yu, Z., Fang, W., Xie, X., Huang, T., and Masquelier, T.
  (2022{\natexlab{c}}).
\newblock Training spiking neural networks with event-driven backpropagation.
\newblock In \emph{36th Conference on Neural Information Processing Systems
  (NeurIPS 2022)}
\bibAnnoteFile{zhu2022training}

\bibitem[{Zimmer et~al.(2019)Zimmer, Pellegrini, Singh, and
  Masquelier}]{2019Technical}
Zimmer, R., Pellegrini, T., Singh, S.~F., and Masquelier, T. (2019).
\newblock Technical report: supervised training of convolutional spiking neural
  networks with pytorch.
\newblock \emph{arXiv preprint arXiv:1911.10124}
\bibAnnoteFile{2019Technical}

\bibitem[{Zou et~al.(2023)Zou, Mu, Zuo, Wang, and Cheng}]{zou2023eventbased}
Zou, S., Mu, Y., Zuo, X., Wang, S., and Cheng, L. (2023).
\newblock Event-based human pose tracking by spiking spatiotemporal
  transformer.
\newblock \emph{arXiv preprint arXiv:2303.09681}
\bibAnnoteFile{zou2023eventbased}

\end{thebibliography}

%%% Make sure to upload the bib file along with the tex file and PDF
%%% Please see the test.bib file for some examples of references

%%% If you don't add the figures in the LaTeX files, please upload them when submitting the article.
%%% Frontiers will add the figures at the end of the provisional pdf automatically
%%% The use of LaTeX coding to draw Diagrams/Figures/Structures should be avoided. They should be external callouts including graphics.

\end{document}